\ifdefined\pdfminorversion
\fi
\documentclass{article} 
\usepackage[T1]{fontenc}
\PassOptionsToPackage{table}{xcolor}
\usepackage{iclr2027_conference,times}

\usepackage{amsmath,amsfonts,bm}

\def\eqref#1{equation~\ref{#1}}

\def\1{\bm{1}}

\DeclareMathAlphabet{\mathsfit}{\encodingdefault}{\sfdefault}{m}{sl}
\SetMathAlphabet{\mathsfit}{bold}{\encodingdefault}{\sfdefault}{bx}{n}

\usepackage{hyperref}
\usepackage{url}
\usepackage{booktabs}      
\usepackage{graphicx}      
\usepackage{amsmath,amssymb}
\usepackage{multirow}
\usepackage{subcaption}
\usepackage{xcolor}
\usepackage{enumitem}
\usepackage{xspace}
\newcommand{\method}{V-JEPA Hand\xspace}

\title{\method: Simple, Accurate and Fast 3D Hand and Camera Trajectory Recovery from Egocentric Videos}

\author{Ruicheng Li$^{1,*}$ \enspace Qixiu Li$^{1,*}$ \enspace Yu Deng$^{2}$ \enspace Sicheng Xu$^{2}$ \enspace Jiaolong Yang$^{2}$ \enspace Baining Guo$^{2}$\\[3pt]
\normalfont $^{1}$Tsinghua University \quad $^{2}$Microsoft Research}

\iclrfinalcopy 

\begin{document}

\maketitle
\ificlrfinal
\begingroup
\renewcommand{\thefootnote}{\fnsymbol{footnote}}
\footnotetext[1]{Intern at Microsoft Research.}
\endgroup
\fi
\fancyhead{}
\renewcommand{\headrulewidth}{0pt}
\vspace{-0.2in}
\noindent\begin{minipage}{\linewidth}
\centering
\includegraphics[width=\linewidth,trim=8bp 6bp 2bp 13bp,clip]{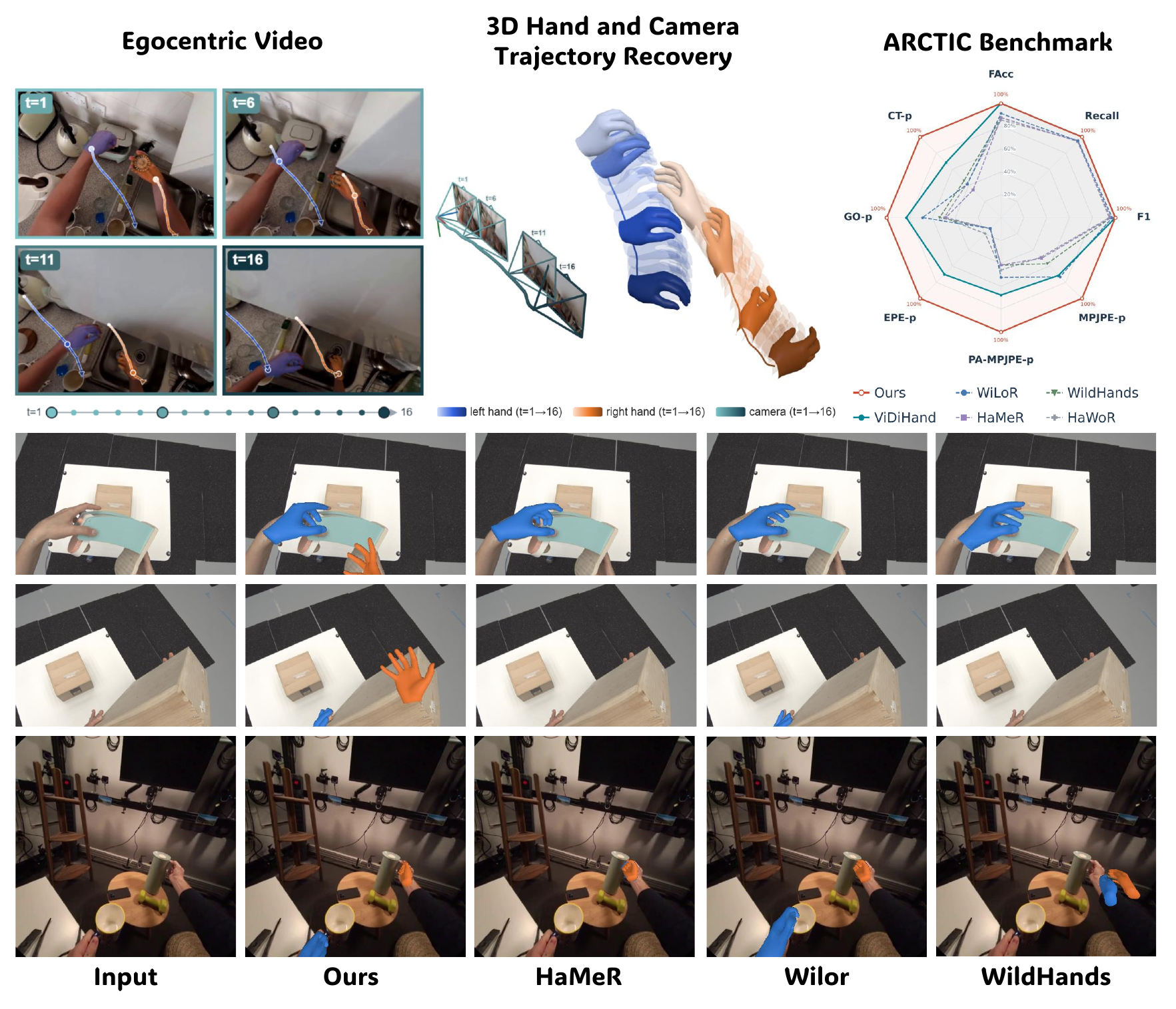}
\captionsetup{font=small,skip=10pt,hypcap=false}
\captionof{figure}{\textbf{\method: joint hand and camera trajectory recovery from egocentric video.}
\textbf{Top left and centre:} From a video clip, \method jointly recovers bilateral 3D hand motion and camera motion in a single forward pass. Blue, orange and teal denote the left hand, right hand and camera, respectively; colour intensity indicates time.
\textbf{Top right:} ARCTIC detection and coverage-aware reconstruction results across eight metrics, normalized to our result (100\%); larger is better.
\textbf{Bottom:} Qualitative comparisons with HaMeR, WiLoR and WildHands, illustrating differences in hand pose accuracy, missed hands and spurious predictions.}
\label{fig:teaser}
\end{minipage}
\par
\vspace{8pt}

\vspace{-6pt}
\begin{abstract}
\vspace{-8pt}
Egocentric video supports diverse applications, including robot learning,
augmented reality, and human activity understanding.
Recovering metric 3D hand trajectories from such video requires estimating
both hand motion and camera egomotion to express the hands in a consistent
reference frame.
Many existing pipelines recover hand and camera motion in separate stages,
combining hand detection, hand reconstruction, camera tracking, and
trajectory refinement.
We present \method, a simple feed-forward video model that jointly
reconstructs bilateral hand motion and camera motion from egocentric clips.
It uses a self-supervised V-JEPA~2.1 encoder to extract spatio-temporal
features from full-frame video and a query-based predictor to decode these
features into per-frame hand poses, hand presence, and camera motion.
We use the predicted camera motion to express the hand trajectories in a
common world frame.
The model accepts known focal lengths or predicts them when unavailable.
It reconstructs each clip in a single forward pass.
We train both components end to end in a single stage on a mixture of
diverse pseudo-labelled videos and high-precision 3D annotations.
A single model achieves the lowest coverage-aware mean per-joint
position error (MPJPE) among comparable baselines on ARCTIC, HOT3D,
and EgoStandard. It also yields lower world-frame trajectory errors than the compared
SLAM-based pipelines. Ablations support the benefits of video pretraining, temporal context,
and mixed supervision.
\method achieves a forward-pass throughput of $93.4$\,FPS on a single
RTX A6000.
Project page: \url{https://qft-333.github.io/V-JEPA-Hand-Page/}.
\end{abstract}

\vspace{-4pt}
\section{Introduction}
\label{sec:intro}

Egocentric video provides a natural record of human manipulation and supports a wide range of applications. Recent VLA approaches use 3D hand motion from
egocentric demonstrations for pretraining
and robot-policy adaptation~\citep{kareer2025egomimic,kareer2025emergence, luo2025beingh0,yang2025egovla,bi2025hrdt,zheng2026egoscale,li2025vitra}.

Hand motion recovery also supports AR/VR interaction~\citep{han2022umetrack},
first-person activity understanding~\citep{kwon2021h2o}, animation and
digital humans~\citep{romero2017mano}, and wrist-kinematics
assessment~\citep{smeragliuolo2016validation}.
These applications benefit from accurate, temporally coherent 3D hand
motion. Apparent hand motion in egocentric video can result from movement
of the hand, the camera, or both. Recovering metric hand trajectories in a
consistent reference frame therefore requires estimating both hands'
positions, orientations, finger articulation, and presence, together with
camera motion.

Single-image models such as HaMeR and
WiLoR~\citep{pavlakos2024hamer,potamias2024wilor} reconstruct hands from
detected crops. Applied frame by frame, they cannot use neighboring frames
to improve reconstruction under occlusion, motion blur, or intermittent
visibility. Temporal methods add
cross-frame modeling~\citep{zhang2024omnihands}, while world-frame approaches
combine hand reconstruction with camera tracking, motion infilling, or
optimization~\citep{zhang2025hawor,chen2025dynhamr}. More recently, ViDiHand~\citep{vidihand} uses pretrained video diffusion features for
detector-free, full-frame hand reconstruction. Video provides temporal cues for hand reconstruction and scene context for
camera motion.
We further investigate \emph{how effectively a self-supervised video encoder can
support joint hand--camera recovery through end-to-end, single-stage
adaptation}.

Our central finding is that \emph{a simple video encoder--predictor
architecture can achieve accurate and fast joint hand--camera reconstruction}.
We instantiate this design as \method, pairing the self-supervised
V-JEPA~2.1 encoder~\citep{murlabadia2026vjepa21} with a query-based predictor
and fine-tuning both end to end. Cross-frame information is incorporated
within the visual encoder, and hand and camera states are decoded from the
same spatio-temporal features. The predictor jointly estimates
per-frame bilateral hand poses, hand presence, and camera motion; it also
accepts known clip-level focal lengths or predicts them when unavailable.
The entire clip is reconstructed in one forward pass without a separate
hand detector, hand crops, SLAM, or test time optimization. The model thus
retains a simple feedforward structure rather than assembling estimates from
independent hand and camera systems.

Training such a model requires accurate 3D annotations across diverse
capture conditions. ARCTIC~\citep{fan2023arctic} and
HOT3D~\citep{banerjee2024hot3d} provide high-quality hand-motion annotations
using dedicated motion-capture systems, while H2O~\citep{kwon2021h2o} uses
calibrated multi-view RGB-D recordings to annotate bimanual interactions.
These resources provide valuable geometric supervision, but their hardware
and calibration requirements make collection costly to extend across
everyday environments. In contrast, Ego4D~\citep{grauman2022ego4d} and
EPIC-KITCHENS~\citep{damen2018epickitchens} capture diverse, unscripted
activities without dense metric hand-motion labels. Reconstruction offers a
complementary source of supervision: HaWoR~\citep{zhang2025hawor} recovers
world-space hand trajectories, while the VITRA annotation
pipeline~\citep{li2025vitra} combines hand reconstruction with metric camera
tracking to generate frame-aligned hand-action labels at scale. Such
automatically recovered trajectories are useful but imperfect pseudo-labels.
We therefore train in a single stage on pseudo-labelled Ego4D and our
in-house recordings, mixed with a comparatively small pool of precisely
annotated videos from ARCTIC and HOT3D. The pseudo-labelled videos broaden
scene and action coverage, while the high-quality annotations provide
reliable geometric supervision for transfer beyond the capture environments.

We evaluate the same trained model on ARCTIC~\citep{fan2023arctic},
HOT3D~\citep{banerjee2024hot3d}, and EgoStandard~\citep{egostandard}, whose
commercial annotation pipeline combines hardware-assisted capture with
multi-stage optimization. EgoStandard is excluded from our reconstruction
training. Across all three benchmarks,
\method combines high hand-detection coverage with the lowest coverage-aware
MPJPE among the previous baselines. It also yields lower world-frame trajectory errors than the
compared SLAM-based pipelines. Data ablations show that mixed supervision
improves reconstruction on unseen videos, while component ablations support
the value of pretrained features and temporal context. \method achieves
forward-pass throughput of $93.4$\,FPS on a single RTX A6000.

We summarize our contributions as follows:
\begin{itemize}[leftmargin=1.4em,itemsep=2pt,topsep=2pt]
  \item A simple yet effective feed-forward framework that jointly recovers
  bilateral 3D hand motion, hand presence, camera egomotion, and clip-level
  focal lengths from a shared video representation, without separate hand
  detection, SLAM, or test-time optimization.
  \item A single-stage training recipe combining diverse pseudo-labelled
  egocentric video with high-precision annotations, with demonstrated
  benefits for reconstruction on unseen data.
  \item A systematic evaluation of detection coverage, hand pose,
  world-frame trajectories, cross-domain generalization, and inference
  efficiency, together with ablations of key training and modeling choices.
\end{itemize}

\section{Related Work}
\label{sec:related}

\paragraph{Monocular 3D hand reconstruction.}
Estimating 3D hand pose and shape from monocular images has advanced rapidly
with parametric models such as MANO~\citep{romero2017mano}.
Most existing methods detect then regress, cropping
each hand with a 2D detector before predicting mesh or kinematic
parameters~\citep{pavlakos2024hamer,potamias2024wilor,dong2024hamba,chen2022mobrecon,moon2023interwild,park2022handoccnet}.
In egocentric videos, however, this pipeline struggles with frequent occlusions,
motion blur, and boundary truncations, while discarding global context and
inter-hand spatial relations.
To bypass cropping, detector-free alternatives directly regress hand poses from
full frames~\citep{prakash2024wildhands} or leverage video diffusion
features~\citep{vidihand}, though diffusion models incur substantial
computational overhead.
In contrast, \method recovers both hands and their presence probabilities in a
single forward pass via a lightweight query-based decoder, avoiding separate
detectors, crops, or iterative diffusion.

\paragraph{Temporal and world-coordinate hand motion recovery.}
Applying single-frame estimators independently across video frames yields
temporally jittery trajectories. While temporal networks and transformers help
smooth camera-relative predictions~\citep{yang2020seqhand,zhang2024omnihands},
they do not resolve camera egomotion.
To express manipulation in a consistent world frame, current methods decouple
hand estimation from camera tracking by combining per-frame regressors,
external visual SLAM such as DROID-SLAM~\citep{teed2021droid}, and motion
infilling or test-time trajectory
optimization~\citep{zhang2025hawor,chen2025dynhamr,fan2024hold}, mirroring
pipelines in world-grounded body capture~\citep{ye2023slahmr,shin2024wham}.
However, SLAM frequently drifts under aggressive head motion, and multi-stage
optimization introduces latency and cascading errors.
Instead, \method unifies camera-space bilateral hand motion, hand presence,
camera egomotion, and camera intrinsics into a single feedforward prediction,
avoiding external SLAM or test-time optimization while achieving superior
world-frame accuracy.

\paragraph{Self-supervised video representations and foundation models.}
Self-supervised video pretraining provides powerful motion and scene priors.
While masked autoencoders such as MAE~\citep{he2022masked},
VideoMAE~\citep{tong2022videomae}, and VideoMAE~V2~\citep{wang2023videomae_v2}
reconstruct masked pixels, pixel-level objectives often overemphasize
high-frequency texture details over geometric and physical dynamics.
The Joint-Embedding Predictive Architecture (JEPA)
family~\citep{assran2023ijepa,bardes2024vjepa,assran2025vjepa2,murlabadia2026vjepa21}
addresses this by predicting representations in latent space; notably,
V-JEPA~2.1~\citep{murlabadia2026vjepa21} learns rich spatio-temporal features
and dense motion priors.
Although generative video foundation models~\citep{blattmann2023stable} also
encode temporal priors~\citep{vidihand}, their generative nature requires heavy
computation.
In comparison, V-JEPA~2.1 offers a discriminative representation that supports
high-throughput ($91.3$\,FPS) inference and outperforms VideoMAE,
DINOv3~\citep{simeoni2025dinov3}, and Perception Encoder~\citep{bolya2025pe} on
both camera- and world-frame metrics.

\section{Method}
\label{sec:method}

\begin{figure}[t]
  \centering
  \includegraphics[width=\linewidth]{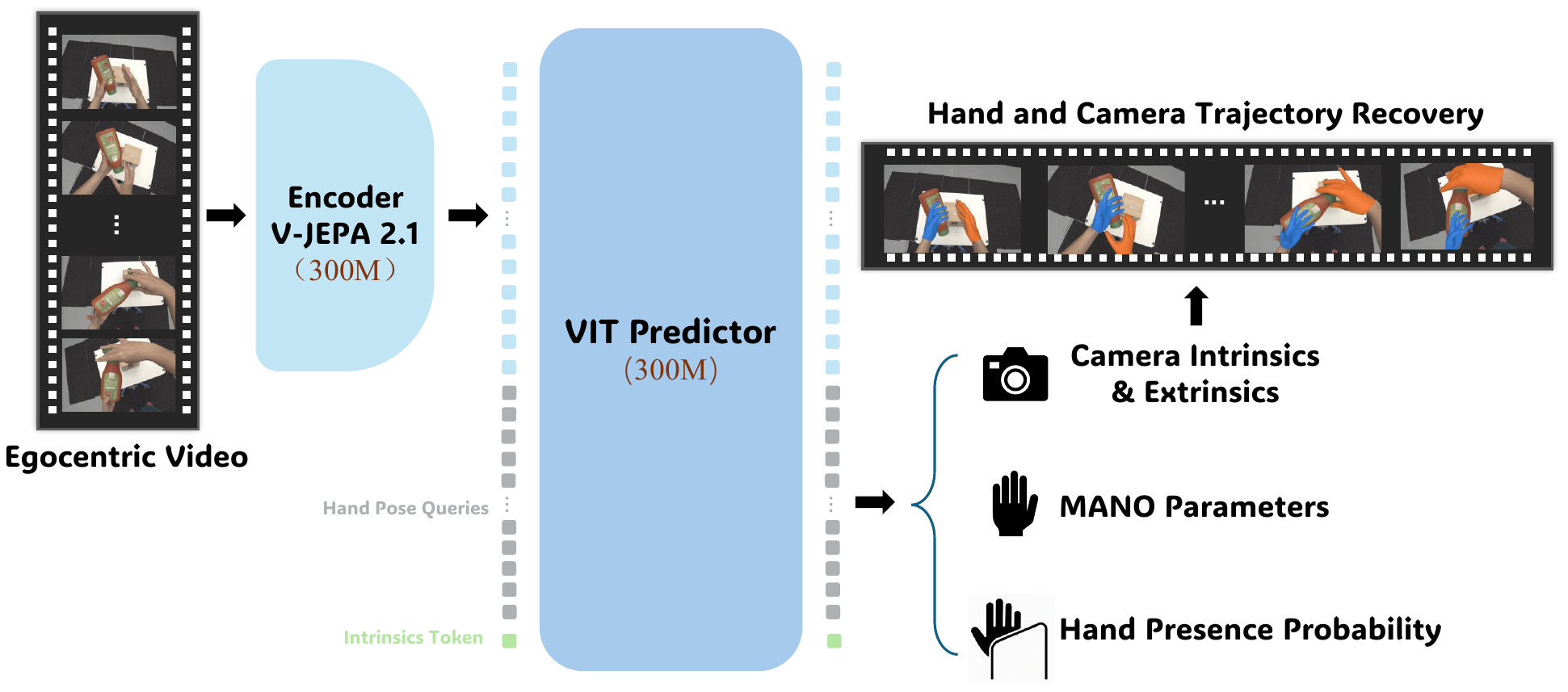}
  \caption{\textbf{Overview of \method.} A V-JEPA~2.1 encoder extracts
  spatio-temporal features from an egocentric video clip. A ViT predictor jointly
  processes these features, one learnable hand-pose query per frame, and an
  intrinsics token. The output queries predict bilateral MANO pose, hand
  presence probability, and camera motion. The entire clip is
  reconstructed in a single forward pass; blue and orange meshes denote the reconstructed
  left and right hands, respectively.}
  \label{fig:arch}
\end{figure}

As illustrated in Figure~\ref{fig:arch}, \method reads a raw video clip and
emits, in a single forward pass, the complete bilateral hand trajectory of that
clip together with the camera motion and focal lengths.
The model has two components: a self-supervised video encoder that turns the
clip into a spatio-temporal token grid (Section~\ref{sec:method:arch}), and a
query-based trajectory predictor that decodes one output token per frame into
the per-frame state defined below. Section~\ref{sec:method:training} describes
our single-stage training with multi-source supervision, combining large-scale
pseudo-labelled video with high-quality data under a shared
training objective.

\subsection{Problem formulation}
\label{sec:method:overview}

The input is a clip of $T$ RGB frames $\mathbf{V}=(\mathbf{I}_1,\dots,
\mathbf{I}_T)$, $\mathbf{I}_t\in\mathbb{R}^{3\times H\times W}$, recorded by a
freely moving egocentric camera. Here $t\in\{1,\dots,T\}$ indexes frames,
and $H$ and $W$ are the image height and width in pixels, respectively.
We write $\mathcal{C}_t$ for the camera
coordinate frame of frame $t$ and index the two hands by
$s\in\{\mathrm{L},\mathrm{R}\}$, where $\mathrm{L}$ and $\mathrm{R}$ denote the
camera wearer's left and right hand. Hats denote model predictions.

\paragraph{Model output.}
The model returns a per-frame hand--camera trajectory together with clip-level
camera focal lengths,
\begin{equation}
  \label{eq:output}
  \hat{\mathbf{Y}}=\big(\hat{\mathbf{y}}_1,\dots,\hat{\mathbf{y}}_T\big)
  \in\mathbb{R}^{T\times119},
  \qquad
  \hat{\mathbf{y}}_t=\big(\,\underbrace{\hat{\mathbf{x}}_t^{\mathrm{L}},\;
  \hat{\mathbf{x}}_t^{\mathrm{R}}}_{110},\;
  \underbrace{\hat{\mathbf{g}}_t}_{9}\,\big),
  \qquad
  \hat{\mathbf{k}}\in\mathbb{R}^{2},
\end{equation}
where $\hat{\mathbf{Y}}$ stacks the per-frame state vectors
$\hat{\mathbf{y}}_t$. Each $\hat{\mathbf{x}}_t^{s}\in\mathbb{R}^{55}$ is the state of hand $s$,
$\hat{\mathbf{g}}_t\in\mathbb{R}^{9}$ is the camera motion at frame $t$, and
$\hat{\mathbf{k}}$ estimates the clip-level focal-length vector
$\mathbf{k}=(f_x,\,f_y)$, where $f_x$ and $f_y$ are the horizontal and
vertical focal lengths in pixels at the input resolution.
Our notation retains only these two varying intrinsic parameters.
The per-frame states are decoded from the
$T$ query tokens, one row $\hat{\mathbf{y}}_t$ per token. The model accepts
known focal lengths or predicts them from a dedicated token when unavailable
(Section~\ref{sec:method:arch}). The hand and camera motion states are defined
next.

\paragraph{Hand state.}
For frame $t$ and hand $s$ the model predicts
\begin{equation}
  \label{eq:hand_state}
  \hat{\mathbf{x}}_t^{s}
  = \big(\,\hat{\mathbf{d}}_t^{s},\;
           \hat{\mathbf{r}}_t^{s},\;
           \hat{\boldsymbol{\theta}}_t^{s},\;
           \hat{c}_t^{s}\,\big)\in\mathbb{R}^{55},
\end{equation}
whose four blocks are: the wrist translation $\hat{\mathbf{d}}_t^{s}\in
\mathbb{R}^{3}$, i.e.\ the position of the hand root in meters expressed in
$\mathcal{C}_t$; the wrist orientation $\hat{\mathbf{r}}_t^{s}\in\mathbb{R}^{6}$,
a prediction of the first two rows of the rotation matrix
$\mathbf{R}_t^{s}\in\mathbb{R}^{3\times3}$ that maps the
canonical MANO hand frame to $\mathcal{C}_t$; the articulation
$\hat{\boldsymbol{\theta}}_t^{s}\in\mathbb{R}^{45}$, the Euler angles of
the $15$ MANO finger joints, each relative to its parent joint; and the presence
score $\hat{c}_t^{s}\in\mathbb{R}$, a logit that is high when hand $s$ is
actually in frame $t$. The two hands together therefore account for the
\textbf{$110$ hand dimensions} of $\hat{\mathbf{y}}_t$.
We do not predict the MANO shape coefficient vector $\boldsymbol{\beta}$, and all hands are evaluated
at the neutral shape $\boldsymbol{\beta}=\mathbf{0}$.

\paragraph{Camera state.}
The camera state encodes the rigid transform
$\mathbf{T}_{t\rightarrow1}\in SE(3)$ from the current camera frame
$\mathcal{C}_t$ to the first-frame anchor $\mathcal{C}_1$, where $SE(3)$
is the group of 3D rigid transformations:
\begin{equation}
  \label{eq:cam_state}
  \hat{\mathbf{g}}_t=\big(\,\hat{\mathbf{r}}^{\text{cam}}_t,\;
  \hat{\mathbf{u}}_t\,\big)\in\mathbb{R}^{9},
  \qquad
  \mathbf{u}_t = \operatorname{sgn}(\mathbf{a}_t)\odot
  \log\!\big(1+|\mathbf{a}_t|\big),
\end{equation}
where the superscript $\text{cam}$ distinguishes camera rotation from hand
rotation. The vector $\hat{\mathbf{r}}^{\text{cam}}_t\in\mathbb{R}^{6}$ predicts
the first two rows of the camera rotation matrix
$\mathbf{R}^{\text{cam}}_t\in\mathbb{R}^{3\times3}$, which maps vectors from
$\mathcal{C}_t$ to $\mathcal{C}_1$.
The vector $\mathbf{a}_t\in\mathbb{R}^{3}$ is the translation component of
$\mathbf{T}_{t\rightarrow1}$, expressed in $\mathcal{C}_1$ in metres.
We represent this translation by its signed-log encoding
$\mathbf{u}_t\in\mathbb{R}^{3}$, and $\hat{\mathbf{u}}_t$ is the model's
prediction of that encoding, rather than the translation in metres.
Here $\operatorname{sgn}$ is the sign function, $\odot$ denotes element-wise
multiplication, and the absolute value and natural logarithm are also applied
element-wise.

\subsection{Architecture}
\label{sec:method:arch}

\paragraph{Video encoder.}
We initialize the video encoder with self-supervised pretrained
V-JEPA~2.1 ViT-L weights~\citep{murlabadia2026vjepa21} and jointly optimize
all encoder and predictor parameters on the mixed training data.
The encoder cuts the video clip into non-overlapping space--time
patches, each spanning $2$ consecutive frames and a $16{\times}16$ pixel square,
and embeds every patch into one token of dimension $d$, so the clip becomes a sequence
\begin{equation}
  \mathbf{Z}\in\mathbb{R}^{N\times d},
  \qquad
  N=\underbrace{\tfrac{T}{2}}_{\text{time}}\cdot
    \underbrace{\tfrac{H}{16}\cdot\tfrac{W}{16}}_{\text{space}} .
\end{equation}
Here $\mathbf{Z}$ is the encoder token sequence, $N$ is the number of
space--time patches, and $d$ is the token feature dimension.
Each token is located by a rotary encoding of its $(\tau,h,w)$ patch
coordinate, where $\tau$, $h$, and $w$ index the temporal, vertical, and
horizontal positions in the patch grid, respectively.

\paragraph{Query-based trajectory predictor.}
The predictor is a ViT that operates on the concatenation of the encoder grid,
$T$ learnable \emph{query tokens} $\mathbf{Q}\in\mathbb{R}^{T\times d}$, one
per output frame, and a single intrinsics token $\mathbf{q}_{k}\in\mathbb{R}^{d}$,
whose subscript $k$ denotes its association with camera intrinsics, under joint
self-attention:
\begin{equation}
  \big[\tilde{\mathbf{q}}_{k};\,\tilde{\mathbf{Z}};\,\tilde{\mathbf{Q}}\big]
  = \operatorname{ViT}\big([\mathbf{q}_{k};\,\mathbf{Z};\,\mathbf{Q}]\big).
\end{equation}
Here $[\cdot;\cdot]$ denotes concatenation along the token dimension, and
$\operatorname{ViT}$ is the predictor's vision transformer.
Tildes denote the corresponding output tokens:
$\tilde{\mathbf{q}}_k\in\mathbb{R}^{d}$ for intrinsics,
$\tilde{\mathbf{Z}}\in\mathbb{R}^{N\times d}$ for the video features, and
$\tilde{\mathbf{Q}}\in\mathbb{R}^{T\times d}$ for the per-frame queries.
Each query is temporally aligned with its corresponding video frame through
rotary positional encoding. A shared MLP maps each output query to the
$119$-dimensional hand--camera state for that frame.
During training, the intrinsics token is set to an embedding of the
ground-truth intrinsics with probability $0.5$ and a learned token otherwise.
A linear head reads the output token $\tilde{\mathbf{q}}_k$ to predict
$\hat{\mathbf{k}}$, the two focal-length components, with supervision applied only to clips whose intrinsics
were withheld.

\subsection{Training}
\label{sec:method:training}

\paragraph{Single-stage multi-dataset training.}
We train \method end to end in a single stage using four datasets:
Ego4D~\citep{grauman2022ego4d}, our in-house egocentric recordings,
ARCTIC~\citep{fan2023arctic}, and HOT3D~\citep{banerjee2024hot3d}.
We process Ego4D and our in-house recordings with the VITRA annotation
pipeline~\citep{li2025vitra} to obtain a large-scale pool of pseudo-labelled
video spanning diverse scenes, activities, and camera wearers.
We mix this noisy pseudo-labelled data with high-quality data from the
training splits of ARCTIC and HOT3D.
Throughout training, each clip is sampled from the pseudo-labelled pool,
ARCTIC, or HOT3D with probabilities $0.35$, $0.45$, and $0.20$, respectively.

\paragraph{Training objective.}
We supervise hand parameters, 3D joint positions, 2D projections, temporal
smoothness, hand presence, and camera estimates with the total training loss
\begin{equation}
  \label{eq:loss}
  \mathcal{L} =
  \underbrace{\lambda_{d}\mathcal{L}_{d}
            + \lambda_{r}\mathcal{L}_{r}
            + \lambda_{\theta}\mathcal{L}_{\theta}}_{\text{parameters}}
  + \underbrace{\lambda_{j}\mathcal{L}_{j}}_{\text{3D}}
  + \underbrace{\lambda_{\pi}\mathcal{L}_{\pi}}_{\text{2D}}
  + \underbrace{\lambda_{a}\mathcal{L}_{a}}_{\text{jitter}}
  + \underbrace{\lambda_{c}\mathcal{L}_{c}
            + \lambda_{g}\mathcal{L}_{g}
            + \lambda_{k}\mathcal{L}_{k}}_{\text{presence and camera}} .
\end{equation}

Each $\lambda$ coefficient is a nonnegative scalar weight for the loss term
with the same subscript.
The terms $\mathcal{L}_{d}$, $\mathcal{L}_{r}$, and $\mathcal{L}_{\theta}$ supervise
wrist position, wrist orientation, and finger articulation, respectively.
$\mathcal{L}_{j}$ penalizes errors in the $21$ camera-space 3D joints obtained
through differentiable MANO forward kinematics, while $\mathcal{L}_{\pi}$
penalizes errors in their image-plane projections.
$\mathcal{L}_{a}$ penalizes the acceleration of the predicted 3D joints across
consecutive frames, which suppresses jitter without penalizing smooth motion.
$\mathcal{L}_{c}$ supervises whether each hand is present in each frame.
$\mathcal{L}_{g}$ supervises camera rotation and signed-log translation
relative to the first frame, and $\mathcal{L}_{k}$ supervises clip-level
focal lengths when they are withheld from the input; auxiliary implementation
details are given in Appendix~\ref{app:loss_details}.

\section{Experiments}
\label{sec:experiments}

We evaluate \method on egocentric two-hand reconstruction. We first describe
the datasets and evaluation protocol (Section~\ref{sec:exp:setup}), then compare
with prior work (Section~\ref{sec:exp:main}). Our ablations test the training
data, the loss terms, encoder pretraining and temporal context, and encoder choice
(Section~\ref{sec:exp:ablation}). We also report model size and inference cost
(Section~\ref{sec:exp:efficiency}), and show qualitative results
(Section~\ref{sec:exp:qual}).

\subsection{Experimental setup}
\label{sec:exp:setup}

\paragraph{Datasets.}
We evaluate on three egocentric hand benchmarks:
\textbf{ARCTIC} tests reconstruction under severe
occlusion during two-hand object manipulation, while
\textbf{HOT3D} tests robustness to motion blur
from rapid head and hand movements. We additionally evaluate on
\textbf{EgoStandard}, a commercially annotated egocentric dataset spanning
diverse scenes. Its 3D motion annotations are produced by a pipeline that
combines hardware-assisted capture with multi-stage optimization. We exclude
EgoStandard entirely from training and use it to assess generalization to
unseen egocentric data.
The EgoStandard evaluation covers approximately $60$ minutes of video.
For HOT3D, we follow the HaWoR validation split, covering approximately
$49$ minutes of video.
Table~\ref{tab:world} uses the same evaluation splits,
with all world methods rescored on their common true-positive support.
\vspace{-15pt}
\paragraph{Metrics.}
Metrics follow the coverage-aware evaluation
implementation of \citet{vidihand}, which accounts for missed hands.
Metric conventions and table abbreviations are provided in
Appendix~\ref{app:covaware}; evaluation protocols are detailed in
Appendix~\ref{app:baselines}.

\subsection{Comparison with prior work}
\label{sec:exp:main}

Table~\ref{tab:main} compares \method with
WiLoR~\citep{potamias2024wilor}, HaMeR~\citep{pavlakos2024hamer},
WildHands~\citep{prakash2024wildhands},
and HaWoR~\citep{zhang2025hawor} on all three datasets.
All five methods are evaluated on the same splits, and \method uses
the same checkpoint across all three datasets.
We also include published ViDiHand~\citep{vidihand} results: its ARCTIC
scores are directly comparable, while its HOT3D scores use a different split
and serve only as a reference.

\begin{table}[t]
\begin{minipage}{\linewidth}
\centering
\footnotesize
\setlength{\tabcolsep}{1.4pt}
\caption{\textbf{Comparison with prior work.} WiLoR, HaMeR, and WildHands are single-frame methods; HaWoR, ViDiHand, and ours are multi-frame methods. We follow ViDiHand's evaluation protocol, except that pose metrics without the -p suffix are computed on each method's own intersection of predictions and ground truth (true positives). ViDiHand results are taken from its paper~\citep{vidihand}; its HOT3D row (gray) uses a different split and is shown for reference only, without direct comparison or ranking.}
\label{tab:main}
\resizebox{\linewidth}{!}{%
\begin{tabular}{lcccccccccccccc}
\toprule
& \multicolumn{3}{c}{Detection} & \multicolumn{4}{c}{3D Pose} & \multicolumn{6}{c}{Orient. \& Position} & \multicolumn{1}{c}{Temporal} \\
\cmidrule(lr){2-4}\cmidrule(lr){5-8}\cmidrule(lr){9-14}\cmidrule(lr){15-15}
Method & Acc$\uparrow$ & Rec$\uparrow$ & F1$\uparrow$ & MP-p$\downarrow$ & MP$\downarrow$ & PA-p$\downarrow$ & PA$\downarrow$ & EP-p$\downarrow$ & EP$\downarrow$ & GO-p$\downarrow$ & GO$\downarrow$ & CT-p$\downarrow$ & CT$\downarrow$ & Jit$\downarrow$ \\
\midrule
\multicolumn{15}{l}{\emph{ARCTIC}} \\
\midrule
WiLoR & \cellcolor{yellow!21}0.909 & \cellcolor{yellow!6}0.949 & \cellcolor{yellow!26}0.971 & \cellcolor{yellow!36}21.11 & \cellcolor{yellow!54}15.52 & \cellcolor{yellow!20}12.74 & \cellcolor{yellow!53}6.70 & 65.56 & \cellcolor{yellow!2}28.84 & \cellcolor{yellow!32}17.68 & \cellcolor{yellow!55}\textbf{11.74} & \cellcolor{yellow!14}0.077 & \cellcolor{yellow!22}0.050 & 25.97 \\
HaMeR & \cellcolor{yellow!8}0.873 & 0.943 & \cellcolor{yellow!13}0.958 & \cellcolor{yellow!3}30.78 & \cellcolor{yellow!9}25.10 & 16.02 & \cellcolor{yellow!2}9.44 & \cellcolor{yellow!1}64.85 & \cellcolor{yellow!3}28.45 & 25.14 & \cellcolor{yellow!1}18.67 & 0.093 & 0.063 & \cellcolor{yellow!27}14.78 \\
WildHands & \cellcolor{yellow!8}0.874 & \cellcolor{yellow!1}0.944 & \cellcolor{yellow!13}0.958 & \cellcolor{yellow!15}27.07 & \cellcolor{yellow!27}21.29 & 15.99 & 9.55 & 65.44 & 29.58 & \cellcolor{yellow!11}22.54 & \cellcolor{yellow!22}16.05 & \cellcolor{yellow!21}0.070 & \cellcolor{yellow!41}0.039 & \cellcolor{yellow!33}12.28 \\
HaWoR & 0.853 & \cellcolor{yellow!10}0.953 & 0.945 & 31.52 & 26.94 & \cellcolor{yellow!9}14.48 & \cellcolor{yellow!9}9.07 & \cellcolor{yellow!21}44.32 & \cellcolor{yellow!35}16.24 & \cellcolor{yellow!3}24.38 & 18.82 & \cellcolor{yellow!15}0.076 & \cellcolor{yellow!22}0.050 & \cellcolor{yellow!16}19.37 \\
ViDiHand & \cellcolor{yellow!55}\textbf{0.997} & \cellcolor{yellow!55}\textbf{0.999} & \cellcolor{yellow!55}\textbf{0.999} & \cellcolor{yellow!34}21.67 & - & \cellcolor{yellow!37}9.82 & - & \cellcolor{yellow!52}12.41 & - & \cellcolor{yellow!46}14.64 & - & \cellcolor{yellow!41}0.047 & - & \cellcolor{yellow!55}\textbf{3.18} \\
Ours & \cellcolor{yellow!55}0.996 & \cellcolor{yellow!54}0.998 & \cellcolor{yellow!55}\textbf{0.999} & \cellcolor{yellow!55}\textbf{15.59} & \cellcolor{yellow!55}\textbf{15.36} & \cellcolor{yellow!55}\textbf{6.84} & \cellcolor{yellow!55}\textbf{6.59} & \cellcolor{yellow!55}\textbf{8.99} & \cellcolor{yellow!55}\textbf{8.85} & \cellcolor{yellow!55}\textbf{12.46} & \cellcolor{yellow!52}12.18 & \cellcolor{yellow!55}\textbf{0.032} & \cellcolor{yellow!55}\textbf{0.031} & \cellcolor{yellow!48}6.09 \\
\midrule
\multicolumn{15}{l}{\emph{HOT3D}} \\
\midrule
WiLoR & \cellcolor{yellow!39}0.810 & \cellcolor{yellow!40}0.879 & \cellcolor{yellow!46}0.922 & \cellcolor{yellow!34}39.38 & \cellcolor{yellow!38}26.09 & \cellcolor{yellow!40}23.10 & \cellcolor{yellow!40}7.57 & \cellcolor{yellow!42}79.90 & \cellcolor{yellow!26}22.11 & \cellcolor{yellow!38}32.51 & \cellcolor{yellow!38}16.49 & \cellcolor{yellow!36}0.103 & \cellcolor{yellow!25}0.048 & \cellcolor{yellow!23}17.19 \\
HaMeR & \cellcolor{yellow!30}0.719 & \cellcolor{yellow!41}0.880 & \cellcolor{yellow!41}0.886 & \cellcolor{yellow!11}67.09 & 57.96 & \cellcolor{yellow!38}25.09 & \cellcolor{yellow!18}10.26 & \cellcolor{yellow!44}70.66 & \cellcolor{yellow!38}15.40 & \cellcolor{yellow!25}50.01 & \cellcolor{yellow!4}36.35 & \cellcolor{yellow!35}0.106 & \cellcolor{yellow!19}0.055 & \cellcolor{yellow!12}21.09 \\
WildHands & 0.425 & 0.600 & 0.609 & 80.23 & \cellcolor{yellow!18}42.99 & 61.89 & 12.40 & 275.39 & 37.14 & 84.35 & 38.82 & 0.250 & 0.076 & 25.30 \\
HaWoR & \cellcolor{yellow!31}0.733 & \cellcolor{yellow!43}0.895 & \cellcolor{yellow!39}0.875 & \cellcolor{yellow!43}28.79 & \cellcolor{yellow!49}16.71 & \cellcolor{yellow!42}20.68 & \cellcolor{yellow!39}7.64 & \cellcolor{yellow!44}66.60 & \cellcolor{yellow!51}8.16 & \cellcolor{yellow!44}25.01 & \cellcolor{yellow!48}10.50 & \cellcolor{yellow!40}0.086 & \cellcolor{yellow!33}0.039 & \cellcolor{yellow!33}13.41 \\
\textcolor{gray}{ViDiHand} & \textcolor{gray}{0.948} & \textcolor{gray}{0.974} & \textcolor{gray}{0.983} & \textcolor{gray}{21.51} & \textcolor{gray}{-} & \textcolor{gray}{11.38} & \textcolor{gray}{-} & \textcolor{gray}{14.95} & \textcolor{gray}{-} & \textcolor{gray}{15.83} & \textcolor{gray}{-} & \textcolor{gray}{0.040} & \textcolor{gray}{-} & \textcolor{gray}{3.74} \\
Ours & \cellcolor{yellow!55}\textbf{0.971} & \cellcolor{yellow!55}\textbf{0.979} & \cellcolor{yellow!55}\textbf{0.980} & \cellcolor{yellow!55}\textbf{14.42} & \cellcolor{yellow!55}\textbf{12.04} & \cellcolor{yellow!55}\textbf{8.24} & \cellcolor{yellow!55}\textbf{5.72} & \cellcolor{yellow!55}\textbf{16.81} & \cellcolor{yellow!55}\textbf{5.71} & \cellcolor{yellow!55}\textbf{9.90} & \cellcolor{yellow!55}\textbf{6.52} & \cellcolor{yellow!55}\textbf{0.026} & \cellcolor{yellow!55}\textbf{0.014} & \cellcolor{yellow!55}\textbf{5.74} \\
\midrule
\multicolumn{15}{l}{\emph{EgoStandard}} \\
\midrule
WiLoR & \cellcolor{yellow!50}0.949 & \cellcolor{yellow!51}0.971 & \cellcolor{yellow!53}0.982 & \cellcolor{yellow!36}35.19 & \cellcolor{yellow!12}32.16 & \cellcolor{yellow!51}11.11 & \cellcolor{yellow!54}7.35 & \cellcolor{yellow!50}37.09 & \cellcolor{yellow!48}12.74 & \cellcolor{yellow!42}25.88 & \cellcolor{yellow!24}22.13 & \cellcolor{yellow!41}0.089 & \cellcolor{yellow!19}0.080 & \cellcolor{yellow!17}15.53 \\
HaMeR & \cellcolor{yellow!55}0.986 & \cellcolor{yellow!55}0.992 & \cellcolor{yellow!55}0.994 & \cellcolor{yellow!34}37.37 & 36.60 & \cellcolor{yellow!55}8.27 & \cellcolor{yellow!55}\textbf{7.26} & \cellcolor{yellow!54}18.06 & \cellcolor{yellow!49}11.45 & \cellcolor{yellow!41}26.37 & \cellcolor{yellow!16}25.37 & \cellcolor{yellow!39}0.093 & \cellcolor{yellow!9}0.090 & \cellcolor{yellow!6}18.35 \\
WildHands & 0.514 & 0.658 & 0.659 & 69.22 & \cellcolor{yellow!8}33.83 & 55.04 & 12.29 & 325.25 & 53.66 & 74.03 & 31.84 & 0.214 & 0.099 & 20.10 \\
HaWoR & \cellcolor{yellow!43}0.883 & \cellcolor{yellow!40}0.901 & \cellcolor{yellow!45}0.937 & \cellcolor{yellow!38}33.38 & \cellcolor{yellow!36}23.36 & \cellcolor{yellow!41}20.11 & \cellcolor{yellow!40}8.62 & \cellcolor{yellow!41}91.58 & \cellcolor{yellow!53}8.00 & \cellcolor{yellow!41}26.58 & \cellcolor{yellow!48}12.81 & \cellcolor{yellow!41}0.089 & \cellcolor{yellow!40}0.058 & \cellcolor{yellow!22}13.95 \\
Ours & \cellcolor{yellow!55}\textbf{0.990} & \cellcolor{yellow!55}\textbf{0.995} & \cellcolor{yellow!55}\textbf{0.997} & \cellcolor{yellow!55}\textbf{16.97} & \cellcolor{yellow!55}\textbf{16.30} & \cellcolor{yellow!55}\textbf{8.01} & \cellcolor{yellow!55}7.29 & \cellcolor{yellow!55}\textbf{11.10} & \cellcolor{yellow!55}\textbf{6.51} & \cellcolor{yellow!55}\textbf{10.60} & \cellcolor{yellow!55}\textbf{9.84} & \cellcolor{yellow!55}\textbf{0.045} & \cellcolor{yellow!55}\textbf{0.043} & \cellcolor{yellow!55}\textbf{4.88} \\
\bottomrule
\end{tabular}}
\end{minipage}
\par\vspace{0.8em}
\begin{minipage}{\linewidth}
\centering
\footnotesize
\setlength{\tabcolsep}{3pt}
\caption{\textbf{World-coordinate reconstruction.} All methods are evaluated on the same hand-frames, using the intersection of true positives across the four compared methods. W-MPJPE and WA-MPJPE are in mm; RTE is in \%. Lower is better.}
\label{tab:world}
\resizebox{\linewidth}{!}{%
\begin{tabular}{lccccccccc}
\toprule
& \multicolumn{3}{c}{ARCTIC} & \multicolumn{3}{c}{HOT3D} & \multicolumn{3}{c}{EgoStandard} \\
\cmidrule(lr){2-4}\cmidrule(lr){5-7}\cmidrule(lr){8-10}
Method & W-MP$\downarrow$ & WA-MP$\downarrow$ & RTE$\downarrow$ & W-MP$\downarrow$ & WA-MP$\downarrow$ & RTE$\downarrow$ & W-MP$\downarrow$ & WA-MP$\downarrow$ & RTE$\downarrow$ \\
\midrule
WiLoR+SLAM & \cellcolor{yellow!33}63.36 & \cellcolor{yellow!33}34.13 & \cellcolor{yellow!34}29.78 & \cellcolor{yellow!23}74.78 & \cellcolor{yellow!19}28.42 & \cellcolor{yellow!26}40.21 & \cellcolor{yellow!8}90.67 & \cellcolor{yellow!16}41.54 & \cellcolor{yellow!39}26.27 \\
HaMeR+SLAM & \cellcolor{yellow!29}69.05 & \cellcolor{yellow!33}33.78 & \cellcolor{yellow!36}28.21 & 105.10 & 35.08 & \cellcolor{yellow!9}47.31 & 94.86 & \cellcolor{yellow!10}42.74 & \cellcolor{yellow!36}27.16 \\
HaWoR & 106.05 & 55.38 & 48.92 & \cellcolor{yellow!33}63.08 & \cellcolor{yellow!22}27.60 & 51.02 & \cellcolor{yellow!9}90.23 & 45.02 & 35.93 \\
Ours & \cellcolor{yellow!55}\textbf{35.86} & \cellcolor{yellow!55}\textbf{19.72} & \cellcolor{yellow!55}\textbf{17.58} & \cellcolor{yellow!55}\textbf{34.08} & \cellcolor{yellow!55}\textbf{16.18} & \cellcolor{yellow!55}\textbf{28.15} & \cellcolor{yellow!55}\textbf{66.35} & \cellcolor{yellow!55}\textbf{32.91} & \cellcolor{yellow!55}\textbf{22.37} \\
\bottomrule
\end{tabular}}
\end{minipage}
\end{table}
\vspace{-6pt}
\paragraph{From occlusion and blur to unseen scenes.}
Table~\ref{tab:main} shows that \method outperforms prior methods on the
vast majority of comparable metrics across all three benchmarks.
The improvements span hand detection, 3D pose, and camera-space placement.
On ARCTIC, this translates into accurate hand recovery despite
severe occlusion. Crucially, the advantage
extends to EgoStandard's diverse scenes without any training on that
dataset: \method leads in most metrics on expanded EgoStandard, including both
coverage-aware and TP-only MPJPE. Together, these results show that a single video-based model supports both precise
reconstruction on ARCTIC and HOT3D and transfer to unseen egocentric scenes.

\paragraph{World-frame reconstruction through unified prediction.}
Table~\ref{tab:world} shows that the simpler architecture also delivers
competitive world-frame trajectories, where hand reconstruction must be
combined with camera motion. \method leads all three world-frame metrics
on all three datasets.
Because this comparison uses shared true-positive hand-frames, the gains
reflect trajectory accuracy rather than differences in detection coverage.
Unlike the SLAM-based baselines, \method predicts hand and camera motion
together from the same video representation, eliminating the need to
assemble their estimates from separate systems. The results therefore
support unified prediction as an accurate alternative to a multi-stage
SLAM-based reconstruction pipeline on these benchmarks.

\subsection{Ablation studies}
\label{sec:exp:ablation}

Ablations use different training setups from Table~\ref{tab:main}.
For fair comparisons, we change only the component being tested and keep
the remaining training settings the same in each ablation.
See Appendix~\ref{app:ablation_details} for implementation details.

\paragraph{Effect of pseudo-labelled training data.}
\label{sec:exp:abl:data}

Table~\ref{tab:abl_data} compares training on all four datasets with
training on an equal mixture of ARCTIC and HOT3D only, keeping the architecture,
initialization, and training budget fixed. We evaluate both models on
EgoStandard, which neither sees during training.
Adding the pseudo-labelled Ego4D and in-house data helps the model place
hands more accurately in these unseen videos: reprojection and translation
errors fall, along with pose and world-trajectory errors.
The benefit is not simply that the model detects more hands---detection
accuracy and recall are slightly lower, while F1 is comparable.
Instead, the broader training data improves the geometry of the reconstruction
on a benchmark outside either model's training set.

\paragraph{Loss ablation.}
\label{sec:exp:abl:loss}

Table~\ref{tab:abl_loss} removes either the jitter loss or the 2D reprojection
loss while keeping the rest of training unchanged.
Without the jitter loss, predictions vary more from frame to frame even
though individual-frame pose accuracy changes little. This shows why
accurate poses alone do not guarantee a smooth sequence.
Without the reprojection loss, joint-position error improves slightly, but
the hands line up less well with the image and their wrist positions become
less accurate. The two terms therefore serve different purposes: one
discourages temporal jumps, while the other keeps the reconstruction
consistent with the image.

\begin{table}[!t]
\centering
\begin{minipage}{\linewidth}
\centering
\footnotesize
\setlength{\tabcolsep}{1.4pt}
\caption{\textbf{Training-data ablation.} To assess how pseudo-labelled training data affects generalization, we compare models trained with and without it on EgoStandard, a benchmark unseen by either model during training.}
\label{tab:abl_data}
\resizebox{\linewidth}{!}{%
\begin{tabular}{lcccccccccccccccc}
\toprule
& \multicolumn{3}{c}{Detection} & \multicolumn{4}{c}{3D Pose} & \multicolumn{6}{c}{Orient. \& Position} & \multicolumn{1}{c}{Temporal} & \multicolumn{2}{c}{World Frame} \\
\cmidrule(lr){2-4}\cmidrule(lr){5-8}\cmidrule(lr){9-14}\cmidrule(lr){15-15}\cmidrule(lr){16-17}
Method & Acc$\uparrow$ & Rec$\uparrow$ & F1$\uparrow$ & MP-p$\downarrow$ & MP$\downarrow$ & PA-p$\downarrow$ & PA$\downarrow$ & EP-p$\downarrow$ & EP$\downarrow$ & GO-p$\downarrow$ & GO$\downarrow$ & CT-p$\downarrow$ & CT$\downarrow$ & Jit$\downarrow$ & W-MP$\downarrow$ & WA-MP$\downarrow$ \\
\midrule
ARCTIC+HOT3D only & \cellcolor{yellow!55}\textbf{0.996} & \cellcolor{yellow!55}\textbf{0.998} & \textbf{0.998} & 18.68 & 18.41 & 8.69 & 8.39 & 16.66 & 15.01 & 12.05 & 11.75 & 0.088 & 0.087 & 4.99 & 70.95 & 34.30 \\
Ours (4 datasets) & 0.991 & 0.996 & \textbf{0.998} & \cellcolor{yellow!55}\textbf{17.44} & \cellcolor{yellow!55}\textbf{16.92} & \cellcolor{yellow!55}\textbf{8.03} & \cellcolor{yellow!55}\textbf{7.46} & \cellcolor{yellow!55}\textbf{10.28} & \cellcolor{yellow!55}\textbf{6.68} & \cellcolor{yellow!55}\textbf{11.10} & \cellcolor{yellow!55}\textbf{10.50} & \cellcolor{yellow!55}\textbf{0.046} & \cellcolor{yellow!55}\textbf{0.044} & \cellcolor{yellow!55}\textbf{4.94} & \cellcolor{yellow!55}\textbf{64.12} & \cellcolor{yellow!55}\textbf{31.75} \\
\bottomrule
\end{tabular}}
\end{minipage}
\par\medskip

\begin{minipage}{\linewidth}
\centering
\footnotesize
\setlength{\tabcolsep}{1.4pt}
\caption{\textbf{Loss ablation on ARCTIC.}}
\label{tab:abl_loss}
\resizebox{\linewidth}{!}{%
\begin{tabular}{lcccccccccccccccc}
\toprule
& \multicolumn{3}{c}{Detection} & \multicolumn{4}{c}{3D Pose} & \multicolumn{6}{c}{Orient. \& Position} & \multicolumn{1}{c}{Temporal} & \multicolumn{2}{c}{World Frame} \\
\cmidrule(lr){2-4}\cmidrule(lr){5-8}\cmidrule(lr){9-14}\cmidrule(lr){15-15}\cmidrule(lr){16-17}
Method & Acc$\uparrow$ & Rec$\uparrow$ & F1$\uparrow$ & MP-p$\downarrow$ & MP$\downarrow$ & PA-p$\downarrow$ & PA$\downarrow$ & EP-p$\downarrow$ & EP$\downarrow$ & GO-p$\downarrow$ & GO$\downarrow$ & CT-p$\downarrow$ & CT$\downarrow$ & Jit$\downarrow$ & W-MP$\downarrow$ & WA-MP$\downarrow$ \\
\midrule
Full & \textbf{0.997} & \cellcolor{yellow!55}\textbf{0.999} & \textbf{0.999} & \cellcolor{yellow!4}15.79 & 15.63 & \cellcolor{yellow!55}\textbf{7.27} & \cellcolor{yellow!55}\textbf{7.10} & \cellcolor{yellow!55}\textbf{9.91} & \cellcolor{yellow!55}\textbf{9.81} & \cellcolor{yellow!28}11.55 & \cellcolor{yellow!18}11.35 & \cellcolor{yellow!55}\textbf{0.019} & \cellcolor{yellow!55}\textbf{0.018} & \cellcolor{yellow!55}\textbf{6.46} & \cellcolor{yellow!55}\textbf{17.98} & \cellcolor{yellow!55}\textbf{11.98} \\
w/o jitter & \textbf{0.997} & 0.998 & \textbf{0.999} & 15.81 & \cellcolor{yellow!9}15.59 & 7.38 & 7.15 & \cellcolor{yellow!47}10.16 & \cellcolor{yellow!47}10.04 & 11.69 & 11.42 & \cellcolor{yellow!55}\textbf{0.019} & \cellcolor{yellow!55}\textbf{0.018} & 8.29 & 18.33 & 12.24 \\
w/o 2D reproj. & \textbf{0.997} & \cellcolor{yellow!55}\textbf{0.999} & \textbf{0.999} & \cellcolor{yellow!55}\textbf{15.55} & \cellcolor{yellow!55}\textbf{15.38} & \cellcolor{yellow!30}7.32 & \cellcolor{yellow!11}7.14 & 11.54 & 11.45 & \cellcolor{yellow!55}\textbf{11.41} & \cellcolor{yellow!55}\textbf{11.20} & 0.021 & 0.020 & \cellcolor{yellow!47}6.71 & \cellcolor{yellow!41}18.07 & \cellcolor{yellow!21}12.14 \\
\bottomrule
\end{tabular}}
\end{minipage}
\par\medskip

\begin{minipage}{\linewidth}
\centering
\footnotesize
\setlength{\tabcolsep}{1.4pt}
\caption{\textbf{Pretraining and temporal context on ARCTIC and HOT3D.} $^\dagger$Single-frame world errors use ground-truth camera rotation and translation (oracle), rather than predicted cameras, and are excluded from world-column ranking.}
\label{tab:abl_design}
\resizebox{\linewidth}{!}{%
\begin{tabular}{lcccccccccccccccc}
\toprule
& \multicolumn{3}{c}{Detection} & \multicolumn{4}{c}{3D Pose} & \multicolumn{6}{c}{Orient. \& Position} & \multicolumn{1}{c}{Temporal} & \multicolumn{2}{c}{World Frame} \\
\cmidrule(lr){2-4}\cmidrule(lr){5-8}\cmidrule(lr){9-14}\cmidrule(lr){15-15}\cmidrule(lr){16-17}
Method & Acc$\uparrow$ & Rec$\uparrow$ & F1$\uparrow$ & MP-p$\downarrow$ & MP$\downarrow$ & PA-p$\downarrow$ & PA$\downarrow$ & EP-p$\downarrow$ & EP$\downarrow$ & GO-p$\downarrow$ & GO$\downarrow$ & CT-p$\downarrow$ & CT$\downarrow$ & Jit$\downarrow$ & W-MP$\downarrow$ & WA-MP$\downarrow$ \\
\midrule
\multicolumn{17}{l}{\emph{ARCTIC}} \\
\midrule
Ours (16 frames) & \cellcolor{yellow!55}\textbf{0.997} & \cellcolor{yellow!55}\textbf{0.999} & \cellcolor{yellow!55}\textbf{0.999} & \cellcolor{yellow!55}\textbf{15.79} & \cellcolor{yellow!55}\textbf{15.63} & \cellcolor{yellow!55}\textbf{7.27} & \cellcolor{yellow!55}\textbf{7.10} & \cellcolor{yellow!55}\textbf{9.91} & \cellcolor{yellow!55}\textbf{9.81} & \cellcolor{yellow!55}\textbf{11.55} & \cellcolor{yellow!55}\textbf{11.35} & \cellcolor{yellow!55}\textbf{0.019} & \cellcolor{yellow!55}\textbf{0.018} & \cellcolor{yellow!55}\textbf{6.46} & \cellcolor{yellow!55}\textbf{17.98} & \cellcolor{yellow!55}\textbf{11.98} \\
Random init. & 0.993 & 0.996 & 0.998 & 24.74 & 24.32 & 11.00 & 10.52 & 17.07 & 16.34 & 16.89 & 16.38 & 0.032 & 0.030 & \cellcolor{yellow!40}8.74 & 37.82 & 23.20 \\
Frozen encoder & \cellcolor{yellow!55}\textbf{0.997} & \cellcolor{yellow!37}0.998 & \cellcolor{yellow!55}\textbf{0.999} & \cellcolor{yellow!46}17.28 & \cellcolor{yellow!46}17.06 & \cellcolor{yellow!44}7.99 & \cellcolor{yellow!45}7.75 & \cellcolor{yellow!50}10.60 & \cellcolor{yellow!51}10.33 & \cellcolor{yellow!46}12.46 & \cellcolor{yellow!46}12.19 & \cellcolor{yellow!21}0.027 & \cellcolor{yellow!23}0.025 & \cellcolor{yellow!50}7.18 & \cellcolor{yellow!46}21.14 & \cellcolor{yellow!46}13.72 \\
Single frame & \cellcolor{yellow!41}0.996 & \cellcolor{yellow!37}0.998 & \cellcolor{yellow!55}\textbf{0.999} & \cellcolor{yellow!43}17.79 & \cellcolor{yellow!43}17.54 & \cellcolor{yellow!38}8.42 & \cellcolor{yellow!38}8.15 & \cellcolor{yellow!41}11.71 & \cellcolor{yellow!44}11.10 & \cellcolor{yellow!43}12.70 & \cellcolor{yellow!43}12.44 & \cellcolor{yellow!47}0.021 & \cellcolor{yellow!46}0.020 & 14.71 & 21.27$^\dagger$ & 14.06$^\dagger$ \\
\midrule
\multicolumn{17}{l}{\emph{HOT3D}} \\
\midrule
Ours (16 frames) & \cellcolor{yellow!55}\textbf{0.989} & \cellcolor{yellow!50}0.993 & \cellcolor{yellow!55}\textbf{0.997} & \cellcolor{yellow!55}\textbf{13.59} & \cellcolor{yellow!55}\textbf{12.75} & \cellcolor{yellow!55}\textbf{7.46} & \cellcolor{yellow!55}\textbf{6.57} & \cellcolor{yellow!54}8.09 & \cellcolor{yellow!55}\textbf{5.81} & \cellcolor{yellow!55}\textbf{7.98} & \cellcolor{yellow!55}\textbf{7.09} & \cellcolor{yellow!55}\textbf{0.020} & \cellcolor{yellow!55}\textbf{0.017} & \cellcolor{yellow!55}\textbf{6.91} & \cellcolor{yellow!55}\textbf{19.84} & \cellcolor{yellow!55}\textbf{11.49} \\
Random init. & 0.969 & 0.984 & 0.988 & 29.02 & 27.04 & 12.15 & 9.89 & 18.97 & 12.34 & 22.27 & 20.23 & 0.045 & 0.037 & \cellcolor{yellow!41}9.91 & 48.81 & 26.87 \\
Frozen encoder & \cellcolor{yellow!55}\textbf{0.989} & \cellcolor{yellow!55}\textbf{0.994} & \cellcolor{yellow!55}\textbf{0.997} & \cellcolor{yellow!49}15.24 & \cellcolor{yellow!48}14.50 & \cellcolor{yellow!50}7.92 & \cellcolor{yellow!46}7.13 & \cellcolor{yellow!55}\textbf{7.79} & \cellcolor{yellow!50}6.41 & \cellcolor{yellow!49}9.41 & \cellcolor{yellow!49}8.58 & \cellcolor{yellow!46}0.024 & \cellcolor{yellow!44}0.021 & \cellcolor{yellow!50}7.90 & \cellcolor{yellow!48}23.75 & \cellcolor{yellow!48}13.40 \\
Single frame & \cellcolor{yellow!44}0.985 & \cellcolor{yellow!44}0.992 & \cellcolor{yellow!43}0.995 & \cellcolor{yellow!43}16.86 & \cellcolor{yellow!43}15.83 & \cellcolor{yellow!40}8.76 & \cellcolor{yellow!37}7.66 & \cellcolor{yellow!49}9.05 & \cellcolor{yellow!45}7.00 & \cellcolor{yellow!45}10.62 & \cellcolor{yellow!45}9.48 & \cellcolor{yellow!40}0.027 & \cellcolor{yellow!36}0.024 & 18.31 & 25.37$^\dagger$ & 15.63$^\dagger$ \\
\bottomrule
\end{tabular}}
\end{minipage}
\par\medskip

\begin{minipage}{\linewidth}
\centering
\footnotesize
\setlength{\tabcolsep}{1.4pt}
\caption{\textbf{Encoder ablation on ARCTIC and HOT3D.}}
\label{tab:abl_enc}
\resizebox{\linewidth}{!}{%
\begin{tabular}{lcccccccccccccccc}
\toprule
& \multicolumn{3}{c}{Detection} & \multicolumn{4}{c}{3D Pose} & \multicolumn{6}{c}{Orient. \& Position} & \multicolumn{1}{c}{Temporal} & \multicolumn{2}{c}{World Frame} \\
\cmidrule(lr){2-4}\cmidrule(lr){5-8}\cmidrule(lr){9-14}\cmidrule(lr){15-15}\cmidrule(lr){16-17}
Method & Acc$\uparrow$ & Rec$\uparrow$ & F1$\uparrow$ & MP-p$\downarrow$ & MP$\downarrow$ & PA-p$\downarrow$ & PA$\downarrow$ & EP-p$\downarrow$ & EP$\downarrow$ & GO-p$\downarrow$ & GO$\downarrow$ & CT-p$\downarrow$ & CT$\downarrow$ & Jit$\downarrow$ & W-MP$\downarrow$ & WA-MP$\downarrow$ \\
\midrule
\multicolumn{17}{l}{\emph{ARCTIC}} \\
\midrule
V-JEPA 2.1 (ours) & \cellcolor{yellow!55}\textbf{0.997} & \cellcolor{yellow!55}\textbf{0.999} & \cellcolor{yellow!55}\textbf{0.999} & \cellcolor{yellow!55}\textbf{15.79} & \cellcolor{yellow!55}\textbf{15.63} & \cellcolor{yellow!55}\textbf{7.27} & \cellcolor{yellow!55}\textbf{7.10} & \cellcolor{yellow!55}\textbf{9.91} & \cellcolor{yellow!55}\textbf{9.81} & \cellcolor{yellow!55}\textbf{11.55} & \cellcolor{yellow!55}\textbf{11.35} & \cellcolor{yellow!55}\textbf{0.019} & \cellcolor{yellow!55}\textbf{0.018} & \cellcolor{yellow!55}\textbf{6.46} & \cellcolor{yellow!55}\textbf{17.98} & \cellcolor{yellow!55}\textbf{11.98} \\
VideoMAE & 0.978 & 0.988 & 0.994 & \cellcolor{yellow!5}20.26 & \cellcolor{yellow!21}18.62 & 10.34 & \cellcolor{yellow!17}8.57 & 16.65 & \cellcolor{yellow!20}11.56 & 15.00 & \cellcolor{yellow!15}13.28 & 0.038 & 0.030 & \cellcolor{yellow!22}7.96 & \cellcolor{yellow!45}21.69 & \cellcolor{yellow!42}14.42 \\
DINOv3 & \cellcolor{yellow!46}0.994 & \cellcolor{yellow!45}0.997 & \cellcolor{yellow!44}0.998 & \cellcolor{yellow!23}18.66 & \cellcolor{yellow!24}18.32 & \cellcolor{yellow!32}8.58 & \cellcolor{yellow!27}8.20 & \cellcolor{yellow!40}11.79 & \cellcolor{yellow!31}11.02 & \cellcolor{yellow!26}13.34 & \cellcolor{yellow!22}12.95 & \cellcolor{yellow!23}0.030 & \cellcolor{yellow!9}0.028 & \cellcolor{yellow!10}8.47 & \cellcolor{yellow!5}35.76 & \cellcolor{yellow!7}20.73 \\
PE-Spatial & \cellcolor{yellow!46}0.994 & \cellcolor{yellow!45}0.997 & \cellcolor{yellow!44}0.998 & 20.72 & 20.41 & \cellcolor{yellow!14}9.58 & 9.24 & \cellcolor{yellow!28}13.28 & 12.56 & \cellcolor{yellow!10}14.35 & 13.99 & \cellcolor{yellow!35}0.026 & \cellcolor{yellow!28}0.024 & 8.94 & 37.60 & 22.02 \\
\midrule
\multicolumn{17}{l}{\emph{HOT3D}} \\
\midrule
V-JEPA 2.1 (ours) & \cellcolor{yellow!55}\textbf{0.989} & \cellcolor{yellow!55}\textbf{0.993} & \cellcolor{yellow!55}\textbf{0.997} & \cellcolor{yellow!55}\textbf{13.59} & \cellcolor{yellow!55}\textbf{12.75} & \cellcolor{yellow!55}\textbf{7.46} & \cellcolor{yellow!55}\textbf{6.57} & \cellcolor{yellow!55}\textbf{8.09} & \cellcolor{yellow!55}\textbf{5.81} & \cellcolor{yellow!55}\textbf{7.98} & \cellcolor{yellow!55}\textbf{7.09} & \cellcolor{yellow!55}\textbf{0.020} & \cellcolor{yellow!55}\textbf{0.017} & \cellcolor{yellow!55}\textbf{6.91} & \cellcolor{yellow!55}\textbf{19.84} & \cellcolor{yellow!55}\textbf{11.49} \\
VideoMAE & \cellcolor{yellow!21}0.984 & \cellcolor{yellow!28}0.991 & \cellcolor{yellow!28}0.995 & \cellcolor{yellow!28}16.92 & \cellcolor{yellow!29}15.83 & \cellcolor{yellow!24}8.76 & \cellcolor{yellow!23}7.60 & \cellcolor{yellow!29}9.84 & \cellcolor{yellow!32}6.86 & \cellcolor{yellow!28}10.96 & \cellcolor{yellow!29}9.80 & \cellcolor{yellow!28}0.025 & \cellcolor{yellow!31}0.021 & \cellcolor{yellow!28}8.58 & \cellcolor{yellow!50}22.26 & \cellcolor{yellow!50}12.95 \\
DINOv3 & \cellcolor{yellow!14}0.983 & \cellcolor{yellow!14}0.990 & \cellcolor{yellow!14}0.994 & \cellcolor{yellow!21}17.85 & \cellcolor{yellow!22}16.55 & \cellcolor{yellow!18}9.01 & \cellcolor{yellow!22}7.62 & \cellcolor{yellow!20}10.50 & \cellcolor{yellow!21}7.35 & \cellcolor{yellow!21}11.78 & \cellcolor{yellow!23}10.38 & \cellcolor{yellow!16}0.027 & \cellcolor{yellow!18}0.023 & \cellcolor{yellow!6}9.95 & \cellcolor{yellow!6}45.93 & \cellcolor{yellow!7}25.24 \\
PE-Spatial & 0.981 & 0.989 & 0.993 & 20.47 & 19.16 & 9.76 & 8.34 & 11.86 & 8.27 & 14.11 & 12.74 & 0.030 & 0.026 & 10.31 & 49.21 & 27.22 \\
\bottomrule
\end{tabular}}
\end{minipage}
\end{table}

\paragraph{Pretraining and temporal context.}
\label{sec:exp:abl:design}

Table~\ref{tab:abl_design} changes one choice at a time: starting the encoder
from random weights, freezing its pretrained weights, or replacing the video
input with a single frame. The remaining training setup is shared.
Training from scratch gives much less accurate poses and world trajectories
on both benchmarks. Even a frozen pretrained encoder performs substantially
better, and fine-tuning it improves the results further. Thus, pretraining
provides useful features for hand reconstruction, while task-specific training
helps the model make better use of them.
Single-frame predictions remain reasonably accurate in each image, but
are much less smooth over time. Seeing neighboring frames therefore helps
not only with pose accuracy, but also with keeping successive predictions
consistent.

\paragraph{Encoder ablation.}
\label{sec:exp:abl:enc}

Table~\ref{tab:abl_enc} compares four pretrained ViT-L encoders with the
same predictor and fine-tuning recipe. V-JEPA~2.1 gives the most accurate
hand poses and world trajectories, and the smoothest predictions, on both
benchmarks. Video encoders outperform image encoders on world trajectories,
even when the latter provide features from multiple frames.
This suggests that video features help recover motion beyond simply
combining per-frame image features. Differences in pretraining data and
input settings are detailed in Appendix~\ref{app:encoder_setup}.

\subsection{Model size and speed}
\label{sec:exp:efficiency}

\method has $607.1$M parameters and jointly recovers hand and camera
trajectories. Table~\ref{tab:speed} compares single-GPU forward throughput
on the same $100$ real-video clips at batch size one, excluding detection
and postprocessing. Ours uses BF16; baselines use their released FP32
implementations.

\begin{table}[!tbp]
\centering
\captionsetup{skip=3pt}
\caption{\textbf{Forward throughput (FPS$\uparrow$).}}
\label{tab:speed}
\small
\setlength{\tabcolsep}{10pt}
\begin{tabular}{lrrrr}
\toprule
GPU & WiLoR & HaMeR & WildHands & Ours \\
\midrule
RTX A6000 & 16.3 & 19.8 & 21.1 & \textbf{93.4} \\
A100 & 11.6 & 15.4 & 14.5 & \textbf{115.1} \\
\bottomrule
\end{tabular}
\end{table}

\subsection{Qualitative results}
\label{sec:exp:qual}
Figure~\ref{fig:qual} in Appendix~\ref{app:benchmark_qual} shows challenging cases beyond controlled capture
settings, including very low light and outdoor activities. \method recovers
hands that the baselines miss in the darkest example and produces plausible
hand reconstructions in outdoor and cluttered scenes. These examples
provide qualitative evidence of generalization to diverse in-the-wild
conditions. Additional ARCTIC and HOT3D comparisons are provided in
Appendix~\ref{app:benchmark_qual} (Figure~\ref{fig:benchmark_qual}).

\section{Conclusion}
\label{sec:conclusion}

We presented \method, a simple model that jointly reconstructs bilateral
hand motion, camera motion and intrinsics from egocentric video in one forward
pass, without hand detection, SLAM or test-time optimization. Built on
V-JEPA~2.1 and trained in a single stage on mixed pseudo-labelled and
high-quality data, it achieves the lowest coverage-aware MPJPE among reproduced
baselines on ARCTIC, HOT3D and the unseen EgoStandard benchmark, and lower
world-frame MPJPE than SLAM-based pipelines. Ablations show that mixing in
pseudo-labelled video improves generalization, and validate video pretraining,
temporal context and encoder choice. Running at over $90$ frames per second,
\method enables large-scale bilateral hand-motion annotation of egocentric
video to support robot imitation learning in embodied AI.

\bibliographystyle{iclr2027_conference}
\bibliography{references}

@inproceedings{pavlakos2024hamer,
  title     = {Reconstructing Hands in {3D} with Transformers},
  author    = {Pavlakos, Georgios and Shan, Dandan and Radosavovic, Ilija and Kanazawa, Angjoo and Fouhey, David and Malik, Jitendra},
  booktitle = {Proceedings of the IEEE/CVF Conference on Computer Vision and Pattern Recognition (CVPR)},
  year      = {2024}
}

@inproceedings{potamias2024wilor,
  title     = {{WiLoR}: End-to-end 3D Hand Localization and Reconstruction in-the-wild},
  author    = {Potamias, Rolandos Alexandros and Zhang, Jinglei and Deng, Jiankang and Zafeiriou, Stefanos},
  booktitle = {Proceedings of the IEEE/CVF Conference on Computer Vision and Pattern Recognition (CVPR)},
  year      = {2025},
  note      = {arXiv:2409.12259}
}

@inproceedings{dong2024hamba,
  title     = {{Hamba}: Single-view {3D} Hand Reconstruction with Graph-guided Bi-Scanning Mamba},
  author    = {Dong, Haoye and Chharia, Aviral and Gou, Wenbo and Vicente Carrasco, Francisco and De la Torre, Fernando},
  booktitle = {Advances in Neural Information Processing Systems (NeurIPS)},
  year      = {2024}
}

@inproceedings{moon2023interwild,
  title     = {Bringing Inputs to Shared Domains for {3D} Interacting Hands Recovery in the Wild},
  author    = {Moon, Gyeongsik},
  booktitle = {Proceedings of the IEEE/CVF Conference on Computer Vision and Pattern Recognition (CVPR)},
  year      = {2023}
}

@inproceedings{prakash2024wildhands,
  title     = {{3D} Hand Pose Estimation in Everyday Egocentric Images},
  author    = {Prakash, Aditya and Tu, Ruisen and Chang, Matthew and Gupta, Saurabh},
  booktitle = {Proceedings of the European Conference on Computer Vision (ECCV)},
  year      = {2024}
}

@article{zhang2024omnihands,
  title   = {{OmniHands}: Towards Robust {4D} Hand Mesh Recovery via {A} Versatile Transformer},
  author  = {Zhang, Xin and Ren, Yuqi and Kang, Di and Xie, Junyu and others},
  journal = {arXiv preprint arXiv:2405.20330},
  year    = {2024}
}

@article{zhang2025hawor,
  title   = {{HaWoR}: World-Space Hand Motion Reconstruction from Egocentric Videos},
  author  = {Zhang, Jinglei and Deng, Jiankang and Ma, Chao and Potamias, Rolandos Alexandros},
  journal = {arXiv preprint arXiv:2501.02973},
  year    = {2025}
}

@article{chen2025dynhamr,
  title   = {{Dyn-HaMR}: Recovering {4D} Interacting Hand Motion from a Dynamic Camera},
  author  = {Chen, Zhengdi and others},
  journal = {arXiv preprint arXiv:2412.12861},
  year    = {2025}
}

@article{romero2017mano,
  title   = {Embodied Hands: Modeling and Capturing Hands and Bodies Together},
  author  = {Romero, Javier and Tzionas, Dimitrios and Black, Michael J.},
  journal = {ACM Transactions on Graphics (SIGGRAPH Asia)},
  volume  = {36},
  number  = {6},
  year    = {2017}
}

@article{bardes2024vjepa,
  title   = {Revisiting Feature Prediction for Learning Visual Representations from Video},
  author  = {Bardes, Adrien and Garrido, Quentin and Ponce, Jean and Chen, Xinlei and Rabbat, Michael and LeCun, Yann and Assran, Mahmoud and Ballas, Nicolas},
  journal = {Transactions on Machine Learning Research (TMLR)},
  year    = {2024},
  note    = {V-JEPA}
}

@article{assran2025vjepa2,
  title   = {{V-JEPA 2}: Self-Supervised Video Models Enable Understanding, Prediction and Planning},
  author  = {Assran, Mahmoud and Bardes, Adrien and others},
  journal = {arXiv preprint arXiv:2506.09985},
  year    = {2025}
}

@article{murlabadia2026vjepa21,
  title   = {{V-JEPA 2.1}: Unlocking Dense Features in Video Self-Supervised Learning},
  author  = {Mur-Labadia, Lorenzo and Muckley, Matthew and Bar, Amir and Assran, Mido and Sinha, Koustuv and Rabbat, Mike and LeCun, Yann and Ballas, Nicolas and Bardes, Adrien},
  journal = {arXiv preprint arXiv:2603.14482},
  year    = {2026}
}

@inproceedings{assran2023ijepa,
  title     = {Self-Supervised Learning from Images with a Joint-Embedding Predictive Architecture},
  author    = {Assran, Mahmoud and Duval, Quentin and Misra, Ishan and Bojanowski, Piotr and Vincent, Pascal and Rabbat, Michael and LeCun, Yann and Ballas, Nicolas},
  booktitle = {Proceedings of the IEEE/CVF Conference on Computer Vision and Pattern Recognition (CVPR)},
  year      = {2023}
}

@inproceedings{tong2022videomae,
  title     = {{VideoMAE}: Masked Autoencoders are Data-Efficient Learners for Self-Supervised Video Pre-Training},
  author    = {Tong, Zhan and Song, Yibing and Wang, Jue and Wang, Limin},
  booktitle = {Advances in Neural Information Processing Systems (NeurIPS)},
  year      = {2022}
}

@inproceedings{grauman2022ego4d,
  title     = {{Ego4D}: Around the World in 3,000 Hours of Egocentric Video},
  author    = {Grauman, Kristen and Westbury, Andrew and Byrne, Eugene and others},
  booktitle = {Proceedings of the IEEE/CVF Conference on Computer Vision and Pattern Recognition (CVPR)},
  year      = {2022}
}

@inproceedings{fan2023arctic,
  title     = {{ARCTIC}: A Dataset for Dexterous Bimanual Hand-Object Manipulation},
  author    = {Fan, Zicong and Taheri, Omid and Tzionas, Dimitrios and Kocabas, Muhammed and Kaufmann, Manuel and Black, Michael J. and Hilliges, Otmar},
  booktitle = {Proceedings of the IEEE/CVF Conference on Computer Vision and Pattern Recognition (CVPR)},
  year      = {2023}
}

@article{banerjee2024hot3d,
  title   = {{HOT3D}: Hand and Object Tracking in {3D} from Egocentric Multi-View Videos},
  author  = {Banerjee, Prithviraj and Shkodrani, Sindi and Moulon, Pierre and Hampali, Shreyas and Han, Shangchen and Zhang, Fan and Zhang, Linguang and Fountain, Jade and Miller, Edward and Basol, Selen and Newcombe, Richard and Wang, Robert and Engel, Jakob and Hodan, Tomas},
  journal = {arXiv preprint arXiv:2406.09598},
  year    = {2024}
}

@article{vidihand,
  title   = {The Surprising Effectiveness of Video Diffusion Models for Hand
             Motion Reconstruction},
  author  = {Wang, Yuxi and Jin, Chengkai and Liu, Yufei and Ouyang, Wenqi and
             Wei, Tianyi and Zeng, Zhiwei and Huang, Siyuan and Shen, Zhiqi and
             Pan, Xingang},
  journal = {arXiv preprint arXiv:2606.30308},
  year    = {2026}
}

@article{simeoni2025dinov3,
  title   = {{DINOv3}},
  author  = {Sim{\'e}oni, Oriane and Vo, Huy V. and Seitzer, Maximilian and Baldassarre, Federico and Oquab, Maxime and Jose, Cijo and Khalidov, Vasil and Szafraniec, Marc and Yi, Seungeun and Ramamonjisoa, Micha{\"e}l and others},
  journal = {arXiv preprint arXiv:2508.10104},
  year    = {2025}
}

@article{bolya2025pe,
  title   = {Perception Encoder: The best visual embeddings are not at the output of the network},
  author  = {Bolya, Daniel and Huang, Po-Yao and Sun, Peize and Cho, Jang Hyun and Madotto, Andrea and Wei, Chen and Ma, Tengyu and Zhi, Jiale and Rajasegaran, Jathushan and Rasheed, Hanoona and others},
  journal = {arXiv preprint arXiv:2504.13181},
  year    = {2025}
}

@article{luo2025beingh0,
  title={Being-h0: vision-language-action pretraining from large-scale human videos},
  author={Luo, Hao and Feng, Yicheng and Zhang, Wanpeng and Zheng, Sipeng and Wang, Ye and Yuan, Haoqi and Liu, Jiazheng and Xu, Chaoyi and Jin, Qin and Lu, Zongqing},
  journal={arXiv preprint arXiv:2507.15597},
  year={2025}
}

@article{yang2025egovla,
  title={Egovla: Learning vision-language-action models from egocentric human videos},
  author={Yang, Ruihan and Yu, Qinxi and Wu, Yecheng and Yan, Rui and Li, Borui and Cheng, An-Chieh and Zou, Xueyan and Fang, Yunhao and Cheng, Xuxin and Qiu, Ri-Zhao and others},
  journal={arXiv preprint arXiv:2507.12440},
  year={2025}
}

@article{zheng2026egoscale,
  title={Egoscale: Scaling dexterous manipulation with diverse egocentric human data},
  author={Zheng, Ruijie and Niu, Dantong and Xie, Yuqi and Wang, Jing and Xu, Mengda and Jiang, Yunfan and Casta{\~n}eda, Fernando and Hu, Fengyuan and Tan, You Liang and Fu, Letian and others},
  journal={arXiv preprint arXiv:2602.16710},
  year={2026}
}

@article{li2025vitra,
  title={Scalable vision-language-action model pretraining for robotic manipulation with real-life human activity videos},
  author={Li, Qixiu and Deng, Yu and Liang, Yaobo and Luo, Lin and Zhou, Lei and Yao, Chengtang and Zeng, Lingqi and Feng, Zhiyuan and Liang, Huizhi and Xu, Sicheng and others},
  journal={arXiv preprint arXiv:2510.21571},
  year={2025}
}

@inproceedings{han2022umetrack,
  title     = {{UmeTrack}: Unified multi-view end-to-end hand tracking for {VR}},
  author    = {Han, Shangchen and Wu, Po-chen and Zhang, Yubo and Liu, Beibei and Zhang, Linguang and Wang, Zheng and Si, Weiguang and Zhang, Peizhao and Cai, Yujun and Hodan, Tomas and others},
  booktitle = {SIGGRAPH Asia 2022 conference papers},
  pages     = {1--9},
  year      = {2022}
}

@inproceedings{kwon2021h2o,
  title={H2o: Two hands manipulating objects for first person interaction recognition},
  author={Kwon, Taein and Tekin, Bugra and St{\"u}hmer, Jan and Bogo, Federica and Pollefeys, Marc},
  booktitle={2021 IEEE/CVF International Conference on Computer Vision (ICCV)},
  pages={10118--10128},
  year={2021},
  organization={IEEE}
}

@article{smeragliuolo2016validation,
  title={Validation of the Leap Motion Controller using markered motion capture technology},
  author={Smeragliuolo, Anna H and Hill, N Jeremy and Disla, Luis and Putrino, David},
  journal={Journal of biomechanics},
  volume={49},
  number={9},
  pages={1742--1750},
  year={2016},
  publisher={Elsevier}
}

@inproceedings{damen2018epickitchens,
  title     = {Scaling Egocentric Vision: The {EPIC-KITCHENS} Dataset},
  author    = {Damen, Dima and Doughty, Hazel and Farinella, Giovanni Maria and Fidler, Sanja and Furnari, Antonino and Kazakos, Evangelos and Moltisanti, Davide and Munro, Jonathan and Perrett, Toby and Price, Will and Wray, Michael},
  booktitle = {Proceedings of the European Conference on Computer Vision (ECCV)},
  month     = {September},
  year      = {2018}
}

@inproceedings{chen2022mobrecon,
  title     = {{MobRecon}: Mobile-Friendly Hand Mesh Reconstruction from Monocular Image},
  author    = {Chen, Xingyu and Liu, Yufeng and Dong, Yajiao and Zhang, Xiong and Ma, Chongyang and Xiong, Yanmin and Zhang, Yuan and Guo, Xiaoyan},
  booktitle = {Proceedings of the IEEE/CVF Conference on Computer Vision and Pattern Recognition (CVPR)},
  year      = {2022}
}

@inproceedings{park2022handoccnet,
  title     = {{HandOccNet}: Occlusion-Robust {3D} Hand Mesh Estimation Network},
  author    = {Park, JoonKyu and Oh, Yeonguk and Moon, Gyeongsik and Choi, Hongsuk and Lee, Kyoung Mu},
  booktitle = {Proceedings of the IEEE/CVF Conference on Computer Vision and Pattern Recognition (CVPR)},
  year      = {2022}
}

@inproceedings{yang2020seqhand,
  title     = {{SeqHAND}: {RGB}-Sequence-Based {3D} Hand Pose and Shape Estimation},
  author    = {Yang, John and Chang, Hyung Jin and Lee, Seungeui and Kwak, Nojun},
  booktitle = {Proceedings of the European Conference on Computer Vision (ECCV)},
  year      = {2020}
}

@inproceedings{teed2021droid,
  title     = {{DROID-SLAM}: Deep Visual {SLAM} for Monocular, Stereo, and {RGB-D} Cameras},
  author    = {Teed, Zachary and Deng, Jia},
  booktitle = {Advances in Neural Information Processing Systems (NeurIPS)},
  year      = {2021}
}

@inproceedings{fan2024hold,
  title     = {{HOLD}: Category-Agnostic {3D} Reconstruction of Interacting Hands and Objects from Video},
  author    = {Fan, Zicong and Parelli, Maria and Kadoglou, Maria Eleni and Chen, Xu and Kocabas, Muhammed and Black, Michael J. and Hilliges, Otmar},
  booktitle = {Proceedings of the IEEE/CVF Conference on Computer Vision and Pattern Recognition (CVPR)},
  year      = {2024}
}

@inproceedings{shin2024wham,
  title     = {{WHAM}: Reconstructing World-Grounded Humans with Accurate {3D} Motion},
  author    = {Shin, Soyong and Kim, Juyong and Halilaj, Eni and Black, Michael J.},
  booktitle = {Proceedings of the IEEE/CVF Conference on Computer Vision and Pattern Recognition (CVPR)},
  year      = {2024}
}

@inproceedings{ye2023slahmr,
  title     = {Decoupling Human and Camera Motion from Videos in the Wild},
  author    = {Ye, Vickie and Pavlakos, Georgios and Malik, Jitendra and Kanazawa, Angjoo},
  booktitle = {Proceedings of the IEEE/CVF Conference on Computer Vision and Pattern Recognition (CVPR)},
  year      = {2023}
}

@inproceedings{he2022masked,
  title     = {Masked Autoencoders Are Scalable Vision Learners},
  author    = {He, Kaiming and Chen, Xinlei and Xie, Saining and Li, Yanghao and Doll{\'a}r, Piotr and Girshick, Ross},
  booktitle = {Proceedings of the IEEE/CVF Conference on Computer Vision and Pattern Recognition (CVPR)},
  year      = {2022}
}

@inproceedings{wang2023videomae_v2,
  title     = {{VideoMAE V2}: Scaling Video Masked Autoencoders with Dual Masking},
  author    = {Wang, Limin and Huang, Bingkun and Zhao, Zhiyu and Tong, Zhan and He, Yinan and Wang, Yi and Wang, Yali and Qiao, Yu},
  booktitle = {Proceedings of the IEEE/CVF Conference on Computer Vision and Pattern Recognition (CVPR)},
  year      = {2023}
}

@article{blattmann2023stable,
  title   = {Stable Video Diffusion: Scaling Latent Video Diffusion Models to Large Datasets},
  author  = {Blattmann, Andreas and Dockhorn, Tim and Kulal, Sumith and Mendelevitch, Daniel and Kilian, Maciej and Lorenz, Dominik and Levi, Yam and English, Zion and Voleti, Vikram and Letts, Adam and Jampani, Varun and Rombach, Robin},
  journal = {arXiv preprint arXiv:2311.15127},
  year    = {2023}
}

@inproceedings{kareer2025egomimic,
  title={Egomimic: Scaling imitation learning via egocentric video},
  author={Kareer, Simar and Patel, Dhruv and Punamiya, Ryan and Mathur, Pranay and Cheng, Shuo and Wang, Chen and Hoffman, Judy and Xu, Danfei},
  booktitle={2025 IEEE International Conference on Robotics and Automation (ICRA)},
  pages={13226--13233},
  year={2025},
  organization={IEEE}
}

@article{kareer2025emergence,
  title={Emergence of human to robot transfer in vision-language-action models},
  author={Kareer, Simar and Pertsch, Karl and Darpinian, James and Hoffman, Judy and Xu, Danfei and Levine, Sergey and Finn, Chelsea and Nair, Suraj},
  journal={arXiv preprint arXiv:2512.22414},
  year={2025}
}

@inproceedings{bi2025hrdt,
  title={H-rdt: Human manipulation enhanced bimanual robotic manipulation},
  author={Bi, Hongzhe and Wu, Lingxuan and Lin, Tianwei and Tan, Hengkai and Su, Zhizhong and Su, Hang and Zhu, Jun},
  booktitle={Proceedings of the AAAI Conference on Artificial Intelligence},
  volume={40},
  number={22},
  pages={18135--18143},
  year={2026}
}

@misc{egostandard,
  title        = {{EgoStandard}: Hardware-Assisted Egocentric Multi-Modal Hand Motion Benchmark},
  author       = {{Lightwheel AI}},
  howpublished = {\url{https://lightwheel.ai/egosuite}},
  year         = {2025},
}

\clearpage
\appendix

\section*{Appendix}
\subsection*{Contents}

\noindent
\hyperref[app:benchmark_qual]{\ref*{app:benchmark_qual}\quad Additional Qualitative Comparisons}
\dotfill \pageref{app:benchmark_qual}\par
\medskip

\noindent
\hyperref[app:encoder_features]{\ref*{app:encoder_features}\quad Encoder Feature Visualization}
\dotfill \pageref{app:encoder_features}\par
\medskip

\noindent
\hyperref[app:hparams]{\ref*{app:hparams}\quad Hyperparameters}
\dotfill \pageref{app:hparams}\par
\medskip

\noindent
\hyperref[app:loss_details]{\ref*{app:loss_details}\quad Training objective details}
\dotfill \pageref{app:loss_details}\par
\medskip

\noindent
\hyperref[app:baselines]{\ref*{app:baselines}\quad Evaluation protocols and baseline setup}
\dotfill \pageref{app:baselines}\par
\medskip

\noindent
\hyperref[app:covaware]{\ref*{app:covaware}\quad Evaluation metrics and table conventions}
\dotfill \pageref{app:covaware}\par
\medskip

\noindent
\hyperref[app:ablation_details]{\ref*{app:ablation_details}\quad Ablation implementation details}
\dotfill \pageref{app:ablation_details}\par
\medskip

\clearpage

\section{Additional qualitative comparisons}
\label{app:benchmark_qual}

\noindent\begin{minipage}{\linewidth}
  \centering
  \captionsetup{skip=3pt}
  \includegraphics[width=\linewidth,height=0.86\textheight,keepaspectratio]{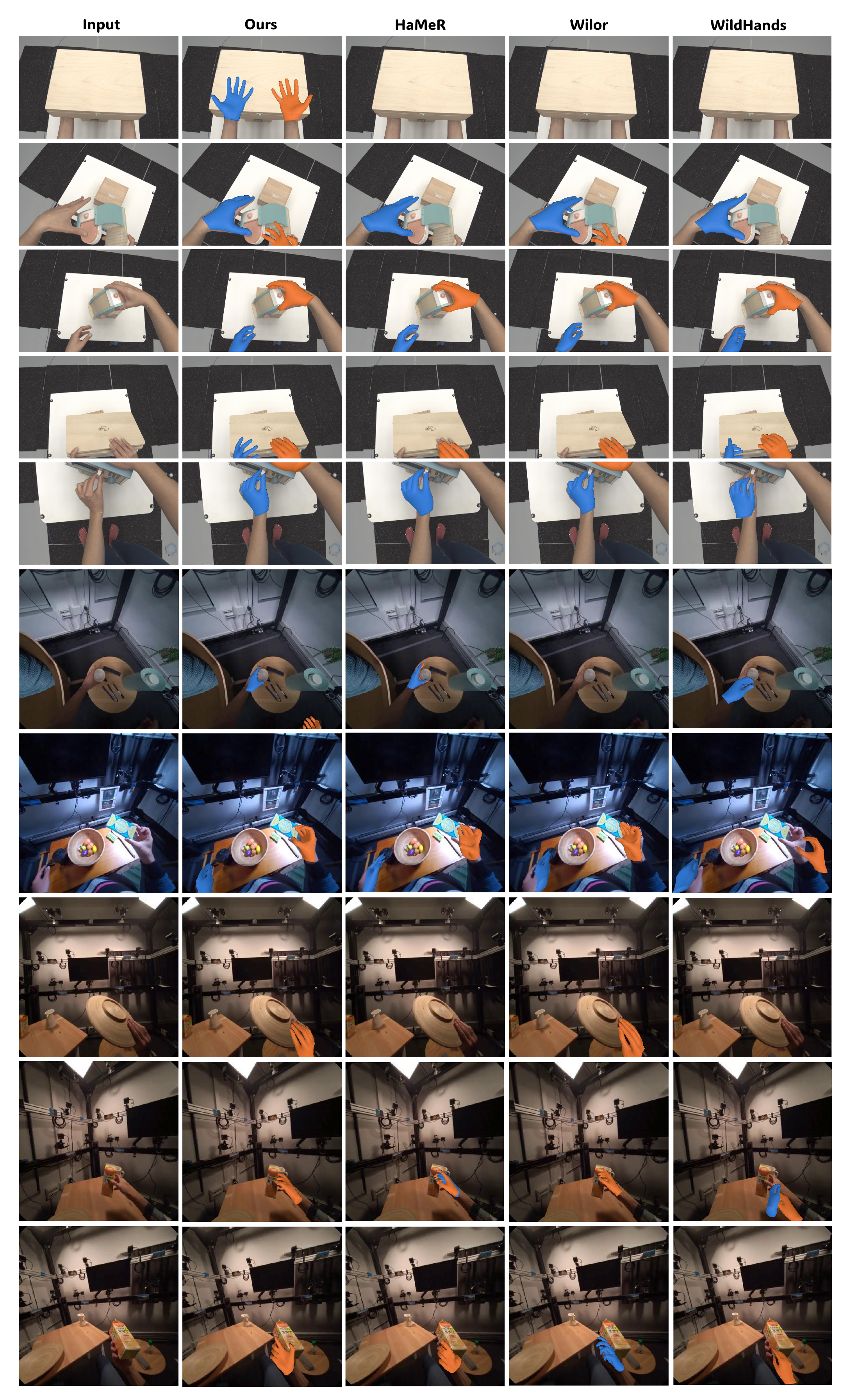}
  \captionof{figure}{\textbf{Qualitative comparisons on ARCTIC and HOT3D.} Each row shows a reference frame and the four methods' reconstructions overlaid on that frame. Blue and orange denote left and right hands, respectively. Examples illustrate missed hands and differences in mesh placement under occlusion and partial visibility.}
  \label{fig:benchmark_qual}
\end{minipage}

\clearpage
\begin{figure}[p]
  \centering
  \includegraphics[width=\linewidth,height=0.73\textheight,keepaspectratio]{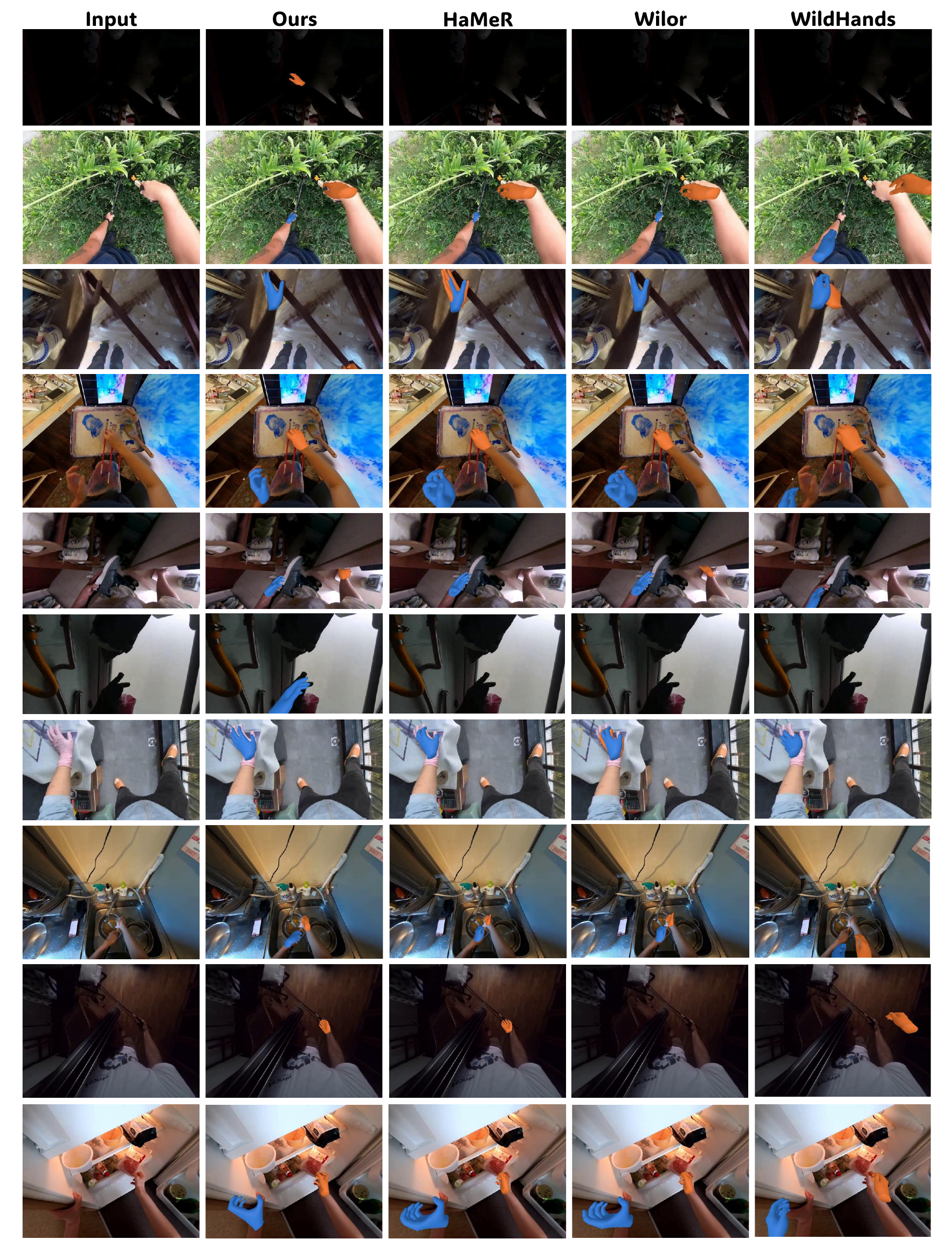}
  \caption{\textbf{Challenging in-the-wild cases.} Comparisons with HaMeR, WiLoR, and WildHands under low light, outdoor backgrounds, and clutter. The first and sixth rows highlight hand reconstructions retained by \method when the baselines miss the hand.}
  \label{fig:qual}
\end{figure}

\clearpage
Figure~\ref{fig:benchmark_qual} compares mesh overlays on ARCTIC and HOT3D
during hand--object interactions. The examples highlight partial visibility,
object occlusion, and hands near the image boundary.
In the ARCTIC box-handling example (Figure~\ref{fig:benchmark_qual}, fourth row
from the top), \method reconstructs both hands near
the lower image boundary, whereas HaMeR and WiLoR omit the left hand.
In the HOT3D plate-holding example (Figure~\ref{fig:benchmark_qual}, eighth row
from the top), \method and WiLoR reconstruct the visible
hand, while HaMeR and WildHands do not show a reconstruction.
Other examples reveal differences in wrist placement and finger articulation
around manipulated objects. These overlays complement the quantitative
results by showing where methods miss hands or produce visibly misaligned
meshes.

Figure~\ref{fig:qual} extends the comparison to challenging in-the-wild
videos, including very low light, outdoor pruning, painting, and household
activities. In the low-light scenes in the first and sixth rows of
Figure~\ref{fig:qual}, \method reconstructs a hand that all three baselines
miss. In the outdoor pruning scene (second row), \method reconstructs both
hands against a background of dense foliage. The painting scene (fourth
row) and the sink scene (eighth row) show hand reconstructions during
object manipulation in cluttered indoor environments.
Across these settings, \method recovers plausible
hand configurations despite substantial changes in lighting, appearance,
and scene layout. These examples provide qualitative evidence of
generalization beyond laboratory settings.

\clearpage
\section{Encoder Feature Visualization}
\label{app:encoder_features}

\begin{figure}[!htbp]
  \centering
  \includegraphics[width=0.92\linewidth]{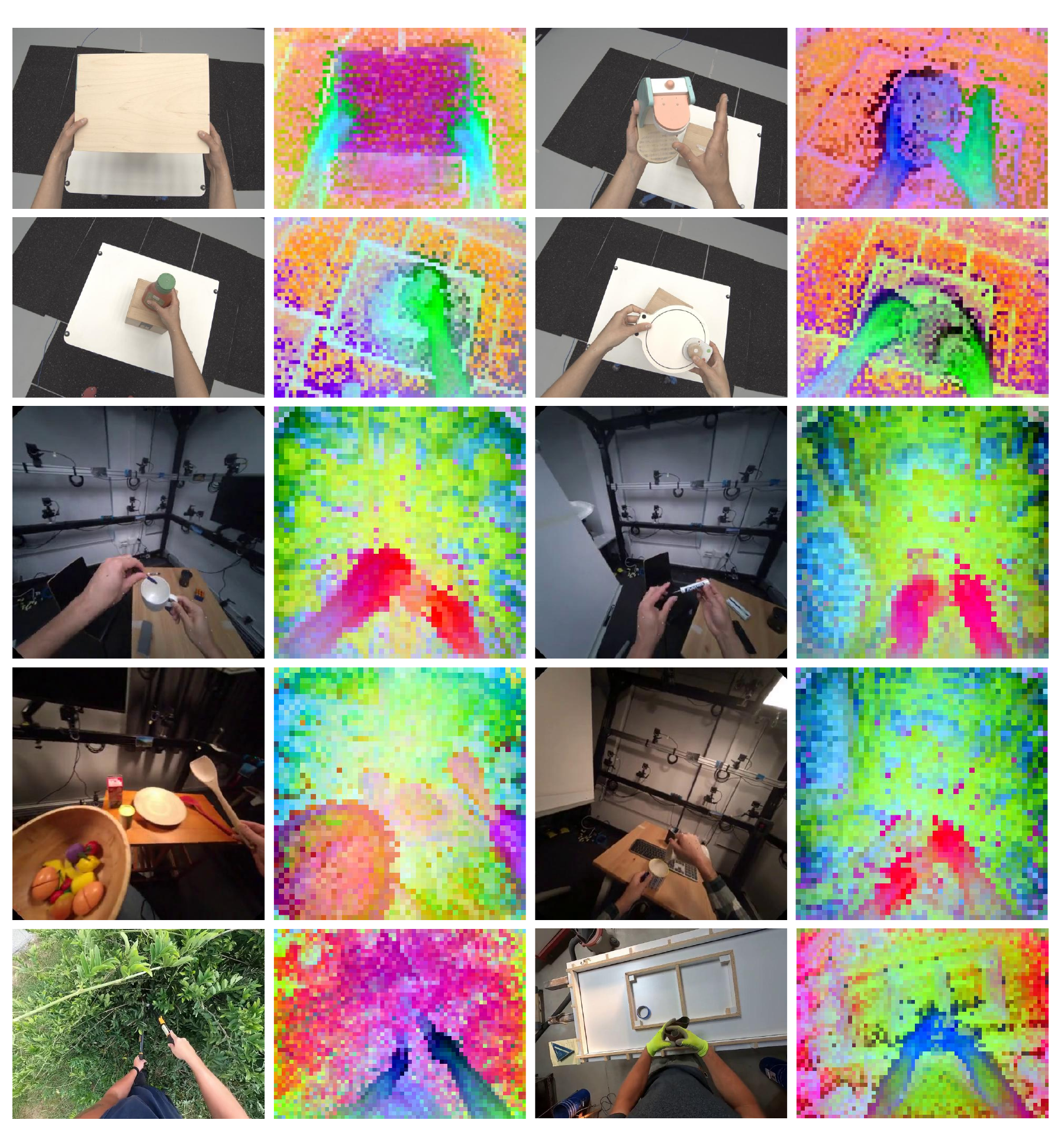}
  \caption{\textbf{Encoder features after task-specific training.} Each input frame is paired with its feature visualization on the right. Coherent patterns corresponding to hands, arms, objects, and scene surfaces suggest that the fine-tuned encoder retains meaningful spatial and semantic structure.}
  \label{fig:encoder_features}
\end{figure}

Figure~\ref{fig:encoder_features} visualizes encoder features after
end-to-end training for hand and camera trajectory recovery.
V-JEPA~2.1 learns dense representations that capture fine-grained spatial
structure as well as global visual information~\citep{murlabadia2026vjepa21}.
Our visualizations suggest that meaningful spatial and semantic structure
remains present after task-specific training: regions corresponding to
hands and forearms form coherent feature patterns, while manipulated
objects and surrounding surfaces often exhibit distinct patterns.
These observations are consistent with retaining semantically structured
features while adapting the encoder to geometric prediction.

\clearpage
\section{Hyperparameters}
\label{app:hparams}

Table~\ref{tab:hparams} lists the architecture, loss, optimization and data
settings of the reported \method model, which is trained in a single stage on
the mixture of pseudo-labelled and high-quality data.

\begin{table}[h]
  \centering
  \caption{\textbf{Hyperparameters.} Model and training settings for \method.
  For smooth-$L_1$ regression, $\delta$ is the absolute error at which the
  penalty changes from quadratic to linear.}
  \label{tab:hparams}
  \small
  \begin{tabular}{lll}
    \toprule
    Group & Hyperparameter & Value \\
    \midrule
    \multirow{4}{*}{Encoder}
      & backbone            & V-JEPA~2.1 ViT-L/384 ( pretrained) \\
      & tubelet / patch     & $2$ (time) / $16{\times}16$ (space) \\
      & tokens per clip     & $8\times24\times24=4608$, dim $1024$ \\
      & fine-tuning         & full (unfrozen) \\
    \midrule
    \multirow{5}{*}{Predictor}
      & type / depth        & ViT-L / $24$ layers \\
      & width / heads       & $1024$ / $16$, MLP ratio $4$ \\
      & query tokens        & $16$ (one per frame) \\
      & attention           & joint self-attention \\
      & stochastic depth    & $0.1$ \\
    \midrule
    \multirow{3}{*}{Targets}
      & end-effector pose   & $9$-dim (3 trans $+$ 6D rot), per hand \\
      & articulation        & $45$-dim MANO Euler angles, per hand \\
      & presence            & $1$ logit per hand per frame \\
    \midrule
    \multirow{7}{*}{Loss}
      & regression          & smooth-$L_1$; weight $1.0$ per term \\
      & \multirow{3}{*}{transition $\delta$} & $0.05$ (wrist and camera rotation) \\
      &                     & $0.03$ (wrist / camera translation, 3D joints) \\
      &                     & $0.02$ (articulation, 2D reproj., focal lengths) \\
      & joint geometry      & neutral-shape MANO FK \\
      & jitter              & $L_1$ joint acceleration, weight $0.2$ \\
      & presence            & BCE, weight $0.5$
       \\
    \midrule
    \multirow{6}{*}{Optimization}
      & optimizer           & AdamW, $\beta{=}(0.9,0.95)$, wd $0.05$ \\
      & lr (predictor)      & $6{\times}10^{-4}$ \\
      & lr (encoder)        & $2{\times}10^{-5}$ \\
      & schedule            & $1{,}800$-step linear warmup $+$ cosine decay, min ratio $5{\times}10^{-4}$ \\
      & training steps      & $30{,}519$ \\
      & effective batch     & $64$ clips ($8$ GPUs $\times$ grad-accum $8$) \\
    \midrule
    \multirow{4}{*}{Data}
      & sources (prob.)     & pseudo-labelled $0.35$, ARCTIC $0.45$, HOT3D $0.20$ \\
      & clip length / size  & $16$ frames / $384{\times}384$ \\
      & augmentation        & frame stride $\{1,2,4,8\}$, horizontal flip, aspect ratio $[1,2]$ \\
      & intrinsics input    & withheld with probability $0.5$ \\
    \bottomrule
  \end{tabular}
\end{table}

\section{Training objective details}
\label{app:loss_details}

We detail the losses in Equation~\ref{eq:loss} using the notation of
Section~\ref{sec:method:overview}: $t$ indexes frames and
$s\in\{\mathrm{L},\mathrm{R}\}$ indexes hands. Hats denote predictions and
unhatted quantities denote targets. We write $\rho_\delta$ for the standard
smooth-$L_1$ loss averaged over coordinates, where $\delta$ is its fixed
quadratic-to-linear transition threshold. For hand geometry,
$\langle\cdot\rangle_{v^s}$ averages over frames with a valid, visible
target for hand $s$ among the video clips processed together in one
forward/backward pass. The two hands' losses are summed.

\paragraph{Hand parameters.}
We directly supervise wrist translation $\mathbf{d}_t^s\in\mathbb{R}^3$
in camera-space metres, wrist orientation $\mathbf{r}_t^s\in\mathbb{R}^6$
in the 6D rotation representation, and finger articulation
$\boldsymbol{\theta}_t^s\in\mathbb{R}^{45}$ as Euler angles in radians:
\begin{equation}
\begin{aligned}
  \mathcal{L}_{d}
    &= \sum_s\langle\rho_{0.03}(\hat{\mathbf{d}}_t^{s},\mathbf{d}_t^{s})\rangle_{v^s}, \\
  \mathcal{L}_{r}
    &= \sum_s\langle\rho_{0.05}(\hat{\mathbf{r}}_t^{s},\mathbf{r}_t^{s})\rangle_{v^s}, \\
  \mathcal{L}_{\theta}
    &= \sum_s\langle\rho_{0.02}(\hat{\boldsymbol{\theta}}_t^{s},
                   \boldsymbol{\theta}_t^{s})\rangle_{v^s}.
\end{aligned}
\end{equation}
The subscripts of $\rho$ give the fixed threshold for each term.

\paragraph{Forward kinematics.}
The differentiable MANO forward-kinematics layer $\mathrm{FK}$ maps the
predicted hand state $\hat{\mathbf{x}}_t^s$ to
$\hat{\mathbf{J}}_t^s=\mathrm{FK}(\hat{\mathbf{x}}_t^s)
\in\mathbb{R}^{21\times3}$, the $21$ camera-space joints in metres,
using neutral hand shape. We supervise these positions with
\begin{equation}
  \mathcal{L}_{j}
  = \sum_s\big\langle\rho_{0.03}\big(\hat{\mathbf{J}}_t^s,
                         \mathbf{J}_t^{s}\big)\big\rangle_{v^s}.
\end{equation}
The target $\mathbf{J}_t^{s}\in\mathbb{R}^{21\times3}$ is the dataset's
measured 3D joints when available and $\mathrm{FK}(\mathbf{x}_t^{s})$
otherwise. This term couples wrist pose and finger articulation through their
effect on joint positions.

\paragraph{Reprojection.}
Let $\pi_{\mathbf{k}}$ project 3D joints to image coordinates using the
ground-truth calibration, with $\mathbf{k}=(f_x,f_y)$ denoting the horizontal
and vertical focal lengths. The reprojection loss encourages image alignment:
\begin{equation}
  \mathcal{L}_{\pi}
  = \sum_s\big\langle\rho_{0.02}\big(
      \pi_{\mathbf{k}}(\hat{\mathbf{J}}_t^s),
      \pi_{\mathbf{k}}(\mathbf{J}_t^{s})\big)\big\rangle_{v^s}.
\end{equation}
Ground-truth intrinsics remain available to this loss even when they are
withheld from the predictor.

\paragraph{Temporal smoothness.}
The jitter loss $\mathcal{L}_a$ is the mean absolute second-order difference
of predicted joints,
\begin{equation}
  \hat{\mathbf{J}}_{t+1}^s-2\hat{\mathbf{J}}_t^s+\hat{\mathbf{J}}_{t-1}^s,
  \qquad t=2,\ldots,T-1,
\end{equation}
where $T$ is the clip length. We average over joint coordinates and triplets
with valid targets in all three sampled frames, then sum over hands.
Constant-velocity motion in camera coordinates incurs no penalty.

\paragraph{Hand presence.}
The presence loss $\mathcal{L}_c$ applies binary cross-entropy to the
predicted logit $\hat{c}_t^s$ and the binary hand-presence target.
It includes both present and absent hands, but excludes unknown targets
and frames outside the anchor recording. We average over valid frames for
each hand and sum the two losses.

\paragraph{Camera motion.}
The camera loss $\mathcal{L}_g$ supervises the transform from each camera
frame to the first-frame anchor. It sums smooth-$L_1$ losses on the
6D rotation and signed-log translation defined in Equation~\ref{eq:cam_state},
with thresholds $0.05$ and $0.03$, respectively. Each block is averaged over
its coordinates and frames with valid camera targets in the anchor recording.

\paragraph{Intrinsics loss.}
The intrinsics loss $\mathcal{L}_k$ supervises the horizontal and vertical
focal lengths $(f_x,f_y)$ using smooth-$L_1$, averaged over the two
components. It is applied only to clips whose intrinsics were withheld
from the predictor.

\paragraph{Loss weights.}
All regression terms have unit weight, including both camera-motion blocks.
The jitter and presence losses have weights $0.2$ and $0.5$, respectively.

\section{Evaluation protocols and baseline setup}
\label{app:baselines}

\paragraph{Baseline implementations.}
We run WiLoR, HaMeR, WildHands and HaWoR for the comparisons in
Table~\ref{tab:main}. ViDiHand's ARCTIC/HOT3D detection, coverage-aware pose and jitter scores are
taken from Table~1 of \citet{vidihand}. Since its pretrained weights are not
publicly available, we cannot reproduce its true-positive-only scores or
evaluate it on EgoStandard.

\paragraph{Hand evaluation.}
We follow ViDiHand's evaluation protocol~\citep{vidihand}.
For ARCTIC, we use the dataset's official validation split; for HOT3D,
we follow the validation split used by HaWoR~\citep{zhang2025hawor}.
The evaluated clips in Table~\ref{tab:main} cover approximately $13$ minutes
of ARCTIC, $49$ minutes of HOT3D and $60$ minutes of EgoStandard.
We use the same trained \method model for all three benchmarks.

\paragraph{World-coordinate comparison.}
\label{app:world_comparison}
Table~\ref{tab:world} uses WiLoR+SLAM, HaMeR+SLAM, HaWoR, and \method.
It uses the same evaluation splits and trained \method model as
Table~\ref{tab:main}.
The +SLAM baselines use the
DROID-SLAM camera implementation distributed with HaWoR, with metric scale estimation;
their camera inference is independent of the hand model.
Unlike the hand-only evaluation in Table~\ref{tab:main},
the world-coordinate evaluation augments WiLoR and HaMeR
with SLAM-based camera estimates.
We first retain clips with completed predictions from all four displayed
methods, then evaluate on their common true-positive hands in each frame.
We evaluate a hand track only if it contains at least three frames with
valid ground truth and true-positive predictions from all four methods.

\paragraph{World-coordinate metrics.}
\label{app:world_metrics}
For the hand tracks selected above, Table~\ref{tab:world} reports joint
position errors (W-MPJPE and WA-MPJPE) and relative wrist trajectory error
(RTE), evaluated over the full clip.
Before measuring errors, we align the predicted hand trajectory to the
ground truth using one rotation and translation for the whole track.
This rigid alignment does not change scale or adjust frames independently.
The metrics differ in how this alignment is estimated and which errors
are measured:
\begin{itemize}[leftmargin=1.4em,itemsep=3pt,topsep=3pt]
  \item \textbf{W-MPJPE: align at the start.}
  We fit the rotation and translation using the $21$ predicted and
  ground-truth hand joints in the first evaluated frame, then apply that
  same transformation to every evaluated frame.
  We report the mean Euclidean joint error over all $21$ joints and all
  evaluated frames, including the first frame, in millimetres.
  This measures how well the reconstruction follows the ground truth
  after the initial alignment.
  \item \textbf{WA-MPJPE: align the whole track.}
  We instead fit a single rotation and translation using the corresponding
  joints from all evaluated frames together.
  We then compute the same mean joint error in millimetres.
  Unlike W-MPJPE, the alignment uses the entire track rather than only
  its first evaluated frame.
  \item \textbf{RTE: normalize wrist error by motion magnitude.}
  Using the whole-track alignment from WA-MPJPE, we compute the mean
  Euclidean distance between predicted and ground-truth wrist positions.
  We divide this error by the largest distance that the ground-truth
  wrist moves from its position in the first evaluated frame, and multiply
  by $100$ to obtain a percentage. The denominator is a displacement
  from the starting position, not the total distance travelled.
  Tracks with this maximum displacement below $1$\,mm are excluded from
  RTE to avoid dividing by a near-zero motion magnitude; they remain
  included in W-MPJPE and WA-MPJPE.
\end{itemize}

\paragraph{Training-data comparison.}
Table~\ref{tab:abl_data} isolates the effect of adding pseudo-labelled
Ego4D and in-house videos to ARCTIC and HOT3D training data.
Both models start from the same pretrained V-JEPA~2.1 encoder and are
trained in a single stage, with the same architecture, training budget,
and HOT3D training-window filter. Ours mixes all four datasets throughout
training; the two-dataset baseline samples ARCTIC and HOT3D equally and
never trains on Ego4D or in-house videos.
Neither model uses the task-specific two-stage recipe employed by the
loss, pretraining and encoder ablations
(Appendix~\ref{app:ablation_setup}).
We evaluate both models on EgoStandard, which is unseen during training,
to test generalization rather than performance on the training datasets.
Hand evaluation uses unsmoothed predictions, MANO-fit targets and projected
MANO mesh boxes for matching. World-coordinate evaluation uses the two
models' common true-positive hand-frames and the full-track metrics
defined in Appendix~\ref{app:world_metrics}.

\paragraph{HOT3D evaluation split.}
The official HOT3D test sequences have no public ground-truth MANO.
Table~\ref{tab:main} uses the HaWoR-aligned held-out split, whereas
ViDiHand uses a random $5\%$ of the validation sequences.
The reported ViDiHand HOT3D scores are therefore shown in gray for
reference only and excluded from ranking.

\section{Evaluation metrics and table conventions}
\label{app:covaware}

\paragraph{Coverage-aware and true-positive-only evaluation.}
We report two complementary views of pose accuracy, following the
coverage-aware evaluation implementation of \citet{vidihand} for metrics
with the $-p$ suffix. The
\emph{coverage-aware} view replaces each unmatched ground-truth hand with a
canonical rest-pose MANO placeholder before averaging. These columns carry the
$\text{-}p$ suffix and reflect both detection coverage and reconstruction
accuracy. The \emph{true-positive-only} view averages pose errors over matched
prediction--ground-truth hands. FAcc, recall and F1 report detection separately.

The main hand comparison and hand-pose ablations report both views.
True-positive-only pose scores are unavailable for ViDiHand and are
marked with dashes, because they cannot be recovered from its published
coverage-aware results.

\paragraph{Table notation and units.}
Table headers abbreviate frame accuracy (FAcc) as \textbf{Acc},
recall as \textbf{Rec}, MPJPE as \textbf{MP}, PA-MPJPE as \textbf{PA},
EPE as \textbf{EP}, and Jitter as \textbf{Jit}.
F1, global-orientation error (\textbf{GO}), and camera-translation error
(\textbf{CT}) retain their short names.
The pairs MP-$p$/MP, PA-$p$/PA, EP-$p$/EP, GO-$p$/GO, and CT-$p$/CT
report coverage-aware and true-positive-only errors, respectively.
In the world comparison and ablations, \textbf{W-MP} and \textbf{WA-MP}
denote W-MPJPE and WA-MPJPE~\citep{zhang2025hawor}.
The world comparison additionally reports relative trajectory error
(\textbf{RTE}) as a percentage; alignment and scoring details are given in
Appendix~\ref{app:world_metrics}.
Across all tables, MP, PA, W-MP, and WA-MP are in millimetres;
EP is in pixels, GO in degrees, CT in metres, and Jit in mm/frame$^2$.
The $-p$ columns use the same units as their unsuffixed counterparts.
Detection, Jit, and world-frame metrics are reported once rather than
as paired columns.
Detection scores lie in $[0,1]$ and are higher-is-better; all other metrics
are lower-is-better. Jitter is true-positive-only in both metric views.
Bold marks column bests where applied, and shading runs from white (worst)
to yellow (best) within each benchmark block. Colours are normalized within
each column and do not indicate the absolute size of an improvement.

\section{Ablation implementation details}
\label{app:ablation_details}

\subsection{Training setup}
\label{app:ablation_setup}
Unlike the single-stage training-data comparison in Table~\ref{tab:abl_data},
the loss, pretraining and encoder ablations use a smaller two-stage recipe.
The first stage trains on half of the Ego4D pseudo-labelled data; the second
fine-tunes the resulting model on ARCTIC and HOT3D.
Each ablation changes one component of this shared setup and is
evaluated after both training phases.
Comparisons within each ablation
isolate the component being studied.

\subsection{World-coordinate evaluation}
\label{app:ablation_world}
Within each component ablation, all variants use the same evaluation
split and the W-MPJPE and WA-MPJPE definitions in
Appendix~\ref{app:world_metrics}.
The single-frame variant is an exception to the camera-motion evaluation:
it cannot estimate camera motion between frames, so its hand predictions
are transformed into world coordinates using ground-truth camera poses
for scoring, without providing these poses to the hand model.
Because this gives it camera information unavailable to the other
variants, its world-coordinate scores are marked with a dagger and
reported for reference only, not included in ranking.
All component ablations share the same HOT3D evaluation split, which
differs from Table~\ref{tab:main}; comparisons therefore focus on variants
within each ablation rather than differences from the main table.

\subsection{Encoder configurations}
\label{app:encoder_setup}
We use the same ViT-L backbone scale and self-attention predictor design,
and match the number of temporal token groups across encoders.
V-JEPA~2.1 and VideoMAE use $16$ input frames with two-frame tubelets,
whereas DINOv3 and PE-Spatial use $8$ independently encoded frames,
yielding eight temporal groups in both cases. This keeps the number of
temporal groups passed to the predictor consistent, although the spatial
grids and input durations differ. Output query counts
match the input frame counts. The configured training image size is $384$
for V-JEPA~2.1 and $224$ for the alternatives; PE-Spatial's positional grid
is interpolated from its pretrained $448$-pixel configuration.
For DINOv3 and PE-Spatial, frames
are processed as a batch of images, prefix tokens are removed, and patch
tokens are concatenated in temporal order for the predictor. Temporal
attention is therefore learned in the predictor rather than applied within
the image backbone.

\end{document}